\documentclass{article}

 \usepackage[preprint]{neurips_2026}

\usepackage[utf8]{inputenc} 
\usepackage[T1]{fontenc}    
\usepackage{hyperref}       
\usepackage{url}            
\usepackage{booktabs}       
\usepackage{amsfonts}       
\usepackage{nicefrac}       
\usepackage{microtype}      
\usepackage{xcolor}         
\usepackage{amsmath}
\usepackage{graphicx}
\usepackage{wrapfig}
\usepackage[]{todonotes}
\usepackage{subcaption}
\usepackage{cleveref}
\newcommand{\best}[1]{\textbf{#1}}
\newcommand{\second}[1]{\underline{#1}}

\newcommand{\fth}{f_\theta}
\newcommand{\E}{\mathbb{E}}
\newcommand{\J}{J}
\newif\ifsubmission

\newcommand{\subvspace}[1]{{\ifsubmission \vspace{#1}\fi}} 

\title{Loss Smoothing for Stable Adaptation Under Distribution Shift}

\author{%
  Darshan Patil$^{1,2,3}$ \quad
  Ekaterina Lobacheva$^{1,2,3}$ \quad
  Razvan Pascanu$^{2}$ \quad
  Sarath Chandar$^{1,2,4,5}$ \\
  $^{1}$Chandar Research Lab \quad
  $^{2}$Mila -- Quebec AI Institute \quad
  $^{3}$Universit\'e de Montr\'eal \\
  $^{4}$Polytechnique Montr\'eal \quad
  $^{5}$Canada CIFAR AI Chair \\
  Correspondence: \texttt{darshan.patil@mila.quebec}
}

\begin{document}

\maketitle

\begin{abstract}
    In settings such as fine-tuning and reinforcement learning, neural networks are often adapted under distribution shift. Standard adaptation methods typically optimize the target objective directly, inducing an abrupt change from the source training objective. This abrupt transition can distort learned representations, including features that may still be useful for the new task.
We investigate whether a more gradual transition can improve adaptation. We propose \emph{loss smoothing}, a simple approach that interpolates between the source and target training objectives at the start of adaptation. This smooth transition helps to preserve useful features from the source distribution while still enabling the model to specialize to the target distribution.
Across controlled supervised shifts, pretrained vision adaptation, offline-to-online and online reinforcement learning, and language model fine-tuning, we find that loss smoothing consistently improves performance, suggesting that smoother objective transitions are a broadly useful tool for model adaptation.
\end{abstract}

\section{Introduction} 
Modern neural networks are rarely trained once and then left unchanged. A pretrained
language model is adapted to instruction-following
data \citep{ouyangTrainingLanguageModels2022, wangHowFarCan2023}; a
classifier trained on one task is updated for the next
\citep{kirkpatrickOvercomingCatastrophicForgetting2017,
sodhaniIntroductionLifelongSupervised2022}; a policy
trained from offline data is fine-tuned through environment interaction
\citep{nairAWACAcceleratingOnline2021, nakamotoCalQLCalibratedOffline2023,
shinOnlinePreTrainingOfflinetoOnline2025}. In each case, adaptation begins from
a model that already contains useful features, but then exposes that model to a
new input data distribution, a new learning signal, or even a new objective. 

The
adaptation problem is therefore not simply to optimize the new objective, but to
do so while making effective use of the representations already present in the model.
This perspective changes the role of stability in adaptation. Stability is often
treated as the opposite of adaptation: a model that stays too close to its
source solution may fail to fit the new task, but the goal is not simply to move
as much as possible. Many shifts contain both \emph{task-shared} components,
such as reusable features, value estimates, behaviors, or representations, and
\emph{task-inconsistent} components that conflict with the new objective.
Successful adaptation should preserve the former while modifying the latter.
For example, an offline RL policy may contain useful exploratory behavior even
when the offline objective is too conservative for online improvement; a
pretrained language model may contain broad general-purpose capabilities even
when instruction tuning pushes it toward a narrow response distribution. Prior
work has observed related brittleness in feature-distorting fine-tuning
\citep{kumar2022finetuning}, offline-to-online RL
\citep{nakamotoCalQLCalibratedOffline2023, shinOnlinePreTrainingOfflinetoOnline2025},
and overtrained language-model fine-tuning
\citep{springerOvertrainedLanguageModels2025b}.

We argue brittle adaptation can arise not only because the target objective is
difficult, but because the transition into it is poorly chosen.
In standard fine-tuning, once the new phase begins, the
source objective is discarded and the model is optimized directly on the target
objective. This hard switch can create large, poorly directed updates exactly
when the model has the least information about which learned representations should be
preserved. 

We propose \emph{loss smoothing}, a simple method for replacing this
discontinuous transition with a smoother objective transition.
At the start of an
adaptation phase, instead of optimizing only the target objective, we optimize a
convex interpolation of the source and target objectives. The interpolation
coefficient is then annealed until ordinary target training is recovered. Early
updates therefore remain close to the source training dynamics, preserving useful
features while the model begins to incorporate target-task information. Later
updates recover the target objective, allowing the model to specialize to the new
task. The motivation is similar to that of continuation methods, curriculum learning,
and neural-network smoothing, which also improve optimization by replacing an
abrupt problem with a smoother path
\citep{allgowerNumericalContinuationMethods1990, bengio2009curriculum,
gulcehreMollifyingNetworks2017}.

Our method is deliberately general. The source and target phases may differ in
their input distributions, their learning signals, or their loss functions. Loss
smoothing only requires an estimate of the previous objective under the current
parameters, which can be obtained through replay data, stored targets, or direct
access to the previous training distribution. This makes it applicable across
settings that are usually studied separately: 
fine-tuning, supervised continual adaptation, and reinforcement learning.

Empirically, we find that smoothing the transition improves adaptation across all
of these settings. In controlled supervised shifts, loss smoothing improves both
test accuracy and trainability, suggesting that preserving useful structure can
make the new task easier to optimize. Additional supervised and pretrained-vision
analyses show that smoothing also reduces transient collapse after task switches
and improves the retention-adaptation trade-off during adaptation. In
offline-to-online reinforcement learning (RL), smoothing provides
a middle ground between conservative methods that remain too close to offline
pretraining and online-only methods that fail when exploration is difficult. In
online RL, smoothing old and new value targets improves performance even on top
of algorithms that already contain stabilizing mechanisms. In language model
fine-tuning, smoothing the transition from pretraining to instruction tuning
reduces overtraining-induced brittleness, showing that catastrophic adaptation
can depend on the path into the downstream objective rather than only on the
pretrained checkpoint or final loss.

Our contributions are:
\begin{itemize}
    \item We formulate adaptation as a transition between training objectives and
    identify the hard switch to the target objective as a source of brittle
    optimization under distribution shift.
    \item We introduce loss smoothing, a lightweight objective interpolation
    method for adaptation across input, output, and objective-function shifts.
    \item We show that this simple path-based intervention improves adaptation
    across continual supervised distribution shifts, pretrained vision adaptation,
    offline-to-online RL, online RL, and LLM fine-tuning.
    \item We provide evidence that preserving reusable features at the start of
    adaptation can lead to stronger downstream specialization later in training.
\end{itemize}

\section{Related Work}
\label{sec:related}

\paragraph{Adaptation under distribution shift.}
Fine-tuning from a pretrained model is a standard approach to adaptation, but
several lines of work show that the resulting update can be brittle under
distribution shift. Fine-tuning can distort pretrained features and hurt
out-of-distribution performance \citep{kumar2022finetuning}, and adaptation can
depend strongly on which parts of the network are allowed to change
\citep{leeSurgicalFineTuningImproves2023a, tomihariUnderstandingLinearProbing2024,
trivediCloserLookModel2022}. Warm-starting a neural network is also not always
equivalent to training from scratch with a better initialization, because the
optimization trajectory itself changes \citep{ashWarmStartingNeuralNetwork2020}.
Our work shares the view that adaptation should preserve useful pretrained
structure, but focuses on a complementary intervention: 
changing the path of intermediate objectives through which optimization moves
from the source objective to the target objective.

\paragraph{Continual Learning.}
Continual learning studies how models adapt across sequences of tasks without
catastrophic forgetting \citep{sodhaniIntroductionLifelongSupervised2022}.
Classical approaches preserve previous knowledge through parameter
regularization \citep{kirkpatrickOvercomingCatastrophicForgetting2017}, replay
\citep{rolnickExperienceReplayContinual2019a}, or distillation from an earlier
model \citep{liLearningForgetting2018}. Dark experience replay further stores
previous model outputs and replays them as targets during later tasks
\citep{buzzegaDarkExperienceGeneral2020}, making it especially close to settings
where loss smoothing estimates an old objective from stored targets. Recent work
has also emphasized transient failures during task changes, including the
stability gap \citep{harunOvercomingStabilityGap2024}, and plasticity loss in
deep and reinforcement learning \citep{dohareLossPlasticityDeep2024,
nikishinPrimacyBiasDeep2022,
tangMitigatingPlasticityLoss2025a,lyle2024disentanglingcausesplasticityloss}. Rather
than
treating stability as a permanent constraint, loss smoothing uses the previous
objective only as a temporary guide at the beginning of adaptation, then anneals
to the target objective. \citet{liuNeuralNetworksLose2026} also study smoothing task
transition boundaries, but rely on distribution or data specific heuristics rather than
a general interpolation of the source and target objectives.

\paragraph{Smooth optimization paths.}
The idea that optimization can be improved by replacing a hard problem with a
sequence of easier or nearby problems appears in continuation methods
\citep{allgowerNumericalContinuationMethods1990}, curriculum learning
\citep{bengio2009curriculum}, and neural-network smoothing or mollification
\citep{gulcehreMollifyingNetworks2017}. Gradual domain adaptation similarly
studies learning through intermediate distributions between a source and target
domain \citep{kumarUnderstandingSelfTrainingGradual2020}. These methods smooth
the data, model, or loss landscape, or assume access to intermediate domains.
Loss smoothing instead addresses a different situation: the learner already has
a source-trained model, and the goal is to avoid a discontinuous switch to a new
objective. The intermediate objectives are constructed by interpolating between
the source and target training losses, making the method applicable across
changes in inputs, learning signals, and objective families.

\paragraph{Offline-to-online reinforcement learning.}
Offline-to-online RL studies how to initialize an agent from a fixed dataset and
then improve it through online interaction. Offline RL methods such as TD3+BC
\citep{fujimotoMinimalistApproachOffline2021, tarasovRevisitingMinimalistApproach2023}
and SPOT \citep{wuSupportedPolicyOptimization2022a} regularize policy learning
to avoid unsupported actions, while offline-to-online methods decide how much
offline data or conservatism should remain during interaction. AWAC uses offline
data to accelerate online RL \citep{nairAWACAcceleratingOnline2021}, Cal-QL
calibrates conservative offline pretraining for efficient online fine-tuning
\citep{nakamotoCalQLCalibratedOffline2023}, and RLPD mixes offline and online
data during online training \citep{ballEfficientOnlineReinforcement2023}. Other
offline-to-online methods modify the pretraining or fine-tuning procedure itself
\citep{shinOnlinePreTrainingOfflinetoOnline2025}.
These methods address the same source-to-target tension that motivates our work:
remaining too close to the offline policy can limit improvement, while switching
too abruptly to unconstrained online RL can destabilize learning. Our
contribution is orthogonal to these algorithmic choices: we smooth the transition
between the offline and online objectives, and in settings where the loss family
changes, we also smooth the objective scale.

\paragraph{Language model fine-tuning.}
Language model fine-tuning also involves an abrupt transition, from broad
pretraining distributions to narrower instruction-tuning or preference-tuning
datasets. Prior work shows that fine-tuning can reduce general-purpose or
out-of-distribution performance \citep{kumar2022finetuning}, and that longer
pretraining can make checkpoints harder to fine-tune through catastrophic
overtraining and progressive sensitivity
\citep{springerOvertrainedLanguageModels2025b}. Source-data mixing is also used
in instruction tuning and alignment; for example, InstructGPT includes a
pretraining-gradient term to reduce capability regressions during RLHF
\citep{ouyangTrainingLanguageModels2022}. Recent work further suggests that
including pretraining data during fine-tuning \citep{kothaReplayingPretrainingData2026}
and specialized fine-tuning data during pretraining
\citep{baekFinetunersFallacyWhen2026} can both improve downstream performance.
We use the setting from \citep{springerOvertrainedLanguageModels2025b} to
test whether the negative effects of overtraining depend not only on the pretraining
checkpoint, but also on the transition path between pretraining and fine-tuning.

\section{Loss Smoothing}
\label{sec:method}
We consider a model $\fth$ with parameters $\theta$ trained through a sequence of
training phases $e = 1,2,\ldots$. 
We view each training phase as inducing a distribution over inputs $u$ and
learning signals $s$. The input $u$ is the object on which the model is evaluated,
such as an image, prompt, state, transition, or trajectory. The signal $s$
specifies how the model should update on that input, such as a label, next-token
target, reward, return, advantage, or value target. Thus, each phase induces an
objective
\begin{equation}
\J_e(\theta)
=
\E_{(u, s) \sim \mathcal{D}_e(\theta)}
\left[
    \ell_e(\theta; u, s)
    \right].
\end{equation}
Our formulation naturally exposes three different sources of distribution shift.
A shift in $u$ corresponds to a change in the distribution of input being
trained on, such as in domain adaptation or an RL policy exploring
different parts of the state space. A shift in $s$ corresponds to a change in the
training signal being provided to the model. This can happen when the target
distribution changes in fine tuning settings, when the reward signal changes in RL, or
even when the target network is updated when bootstrapping in RL. Finally, the last
shift corresponds to a change in $\ell$, i.e. the formulation of the objective.
For example, in offline-to-online RL, the
loss function in the offline phase is different than the loss function in the online
phase. 

Adaptation corresponds to moving from a source objective $\J_s$ to a target
objective $\J_t$. Standard fine-tuning performs an abrupt switch
discarding the source objective once target training begins. Loss smoothing
instead replaces this discontinuous switch with a path of intermediate
objectives:
\[
    \J_{\alpha}(\theta)
    =
    (1-\alpha)\J_s(\theta) + \alpha \J_t(\theta),
    \qquad \alpha \in [0,1].
\]
During adaptation, $\alpha$ is linearly increased over $\tau$ steps according to a schedule
 $\alpha(k) = \min(k/\tau,\,1)$.
At the beginning of adaptation, $\alpha$ is small, so updates remain close to
the source training dynamics. As adaptation proceeds, $\alpha$ increases toward
one, eventually recovering ordinary target training.

Our method is flexible in that it makes no assumptions on what the source
training phase was or how it relates to the current phase. The shifts we
mention can happen independently and even all at once. The intended role of
smoothing is to make the beginning of adaptation informative rather than
destructive. If part of the model implements structure that is shared between
the source and target phases, the source term discourages early updates from
immediately overwriting it. If another part of the model is inconsistent with
the target phase, the target term can still move it. Loss smoothing therefore
acts as a temporary mechanism for separating task-shared components that should
be preserved from task-inconsistent components that should be modified.

This interpretation motivates two design choices. First, the previous phase is
represented through an estimate of the source objective, such as replay data or
stored targets, rather than only through a local penalty on parameters. This
keeps the preservation signal tied to the function or prediction induced by the
source phase even after the parameters move. Second, the source term is annealed
away. If it remains active for the whole target phase, the method becomes closer
to ordinary replay and can prevent specialization to the new task. In our
experiments, we estimate the previous objective using replay buffers
(offline-to-online RL), direct access to the previous data (LLM fine-tuning and
DomainNet ViT adaptation), or a single stored batch of past data (continual supervised
learning and online reinforcement learning).

In \Cref{sec:exp}, we will walk through different case studies of how our method can be
applied to different instantiations of distribution shifts.

\section{Experiments}
\begin{table}[t]
\centering
\small
\setlength{\tabcolsep}{4pt}
\begin{tabular}{@{}p{0.22\textwidth}p{0.27\textwidth}p{0.45\textwidth}@{}}
\toprule
Setting & Source $\to$ target & Shift instantiated \\
\midrule
Supervised Adaptation &
Previous task $\to$ current task &
Permuted inputs isolate a shift in $u$; shuffled labels isolate a shift in $s$,
with cross-entropy loss fixed. \\
Offline-to-Online RL &
Offline $\to$ online &
Transitions, bootstrapped targets, and sometimes the actor objective all change,
corresponding to shifts in $u$, $s$, and $\ell$. \\
LLM Fine-tuning &
Pretraining $\to$ instruction tuning &
Prompts, sequences, and target tokens narrow, shifting both $u$ and $s$ while
retaining a next-token loss form. \\
DomainNet ViT &
ImageNet $\to$ DomainNet &
The visual domain and labels change, while the ImageNet source objective is retained
through replay or smoothing. \\
Online RL &
Previous training phase $\to$ current rollout phase &
The policy changes the visited transitions and bootstrapped value targets,
creating continual shifts in $u$ and $s$. \\
\bottomrule
\end{tabular}
\subvspace{1ex}
\caption{How each experiment instantiates the source-to-target objective
transition from \Cref{sec:method}.}
\label{tab:shift-instantiations}
\end{table}

\label{sec:exp}
We evaluate loss smoothing across case studies that instantiate different parts
of the objective-shift decomposition in \Cref{sec:method}. In each setting, we
identify a source objective $\J_s$, a target objective $\J_t$, and the way the
transition changes inputs $u$, learning signals $s$, or the loss family $\ell$.
\Cref{tab:shift-instantiations} summarizes these instantiations before we give
the experimental details for each case study.
After these case studies, \Cref{sec:analysis} probes the mechanism more directly and
extends the supervised evidence to pretrained vision adaptation.
\subsection{Offline-to-Online RL}

\begin{figure}
    \begin{subfigure}{.24\textwidth}
        \includegraphics[width=\textwidth]{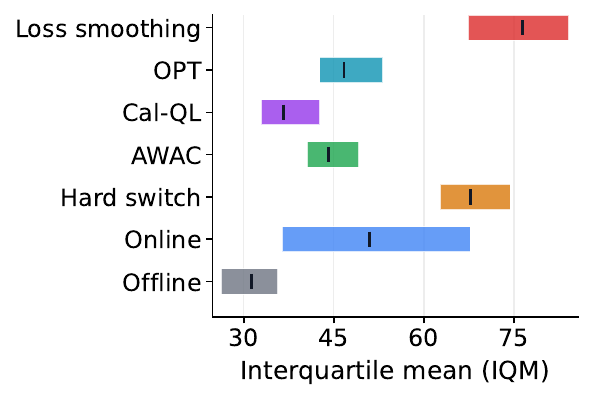}
        \caption{MuJoCo results.}
        \label{fig:iqm_mujoco}
    \end{subfigure}
    \begin{subfigure}{.24\textwidth}
        \includegraphics[width=\textwidth]{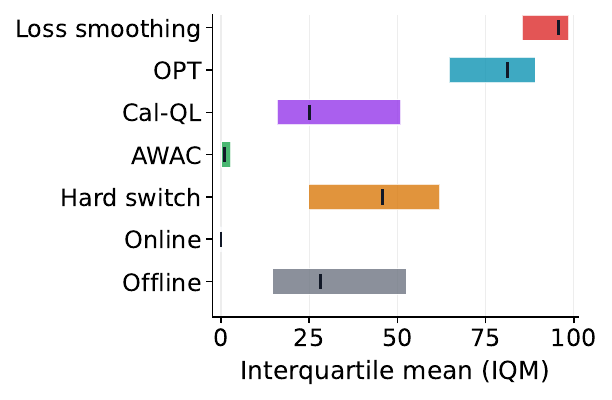}
        \caption{AntMaze results.}
        \label{fig:iqm_antmaze}
    \end{subfigure}
    \begin{subfigure}{.24\textwidth}
        \includegraphics[width=\textwidth]{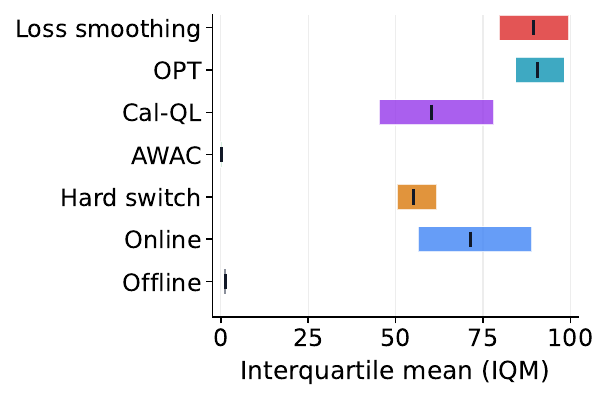}
        \caption{Adroit results.}
        \label{fig:iqm_adroit}
    \end{subfigure}
    \begin{subfigure}{.24\textwidth}
        \includegraphics[width=\textwidth]{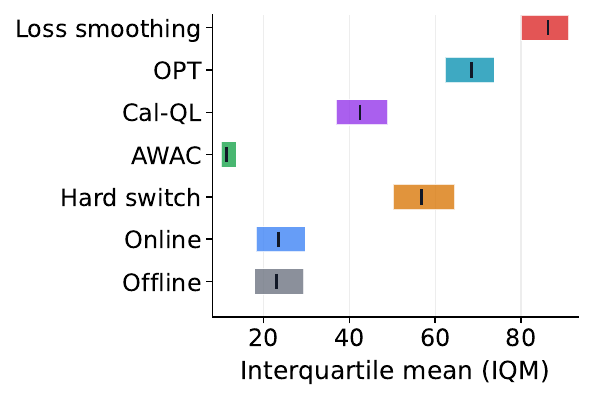}
        \caption{Overall results.}
        \label{fig:iqm_overall}
    \end{subfigure}
    \caption{Suite-level interquartile mean results for the MuJoCo, AntMaze, and
    Adroit offline-to-online RL benchmarks, as well as the overall interquartile mean
    across all environments.
    Values are mean $\pm$ 95\% confidence interval.
    }
    \label{fig:o2o_rliable}
\end{figure}
\begin{wrapfigure}[9]{r}{0.65\textwidth}
    \vspace{-9ex}
    \includegraphics[width=\linewidth]{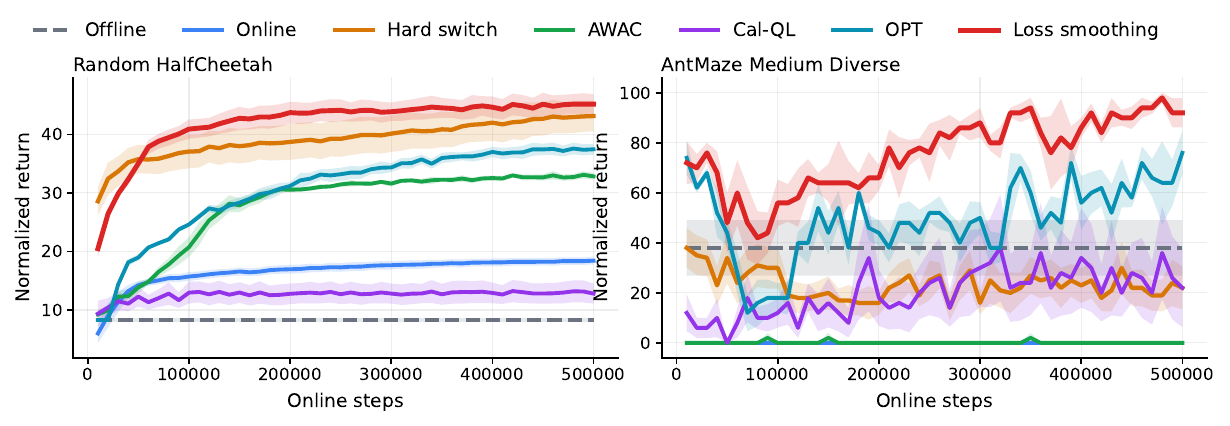}
    \caption{Learning curves on selected environments. Shaded regions denote standard
    error across seeds.
    }
    \label{fig:o2o_sel_curves}
\end{wrapfigure}
Offline-to-online RL is a natural testbed for loss smoothing because adaptation
can change both the data distribution and the optimization objective. An agent is
first pretrained from a fixed offline dataset, and is then fine-tuned with direct
environment interaction. Several prior studies have shown that fine-tuning from offline
trained initializations can result in worse performance than training directly from
scratch \citep{nakamotoCalQLCalibratedOffline2023,
shinOnlinePreTrainingOfflinetoOnline2025}. Policies learned from a fixed dataset can overfit to unsupported
actions and suffer from biased value estimates, yet overly conservative offline training
can limit improvement once interaction begins. 
The key challenge is to properly leverage imperfect offline
knowledge while enabling efficient online exploration and adaptation without
destabilizing performance. 

In the notation of \Cref{sec:method}, $\J_s$ is the offline training objective
estimated from the fixed dataset and $\J_t$ is the online fine-tuning objective;
the transition can change the sampled transitions $u$, the bootstrapped targets
or rewards $s$, and, for TD3+BC to TD3, the actor loss $\ell$.

We evaluate our loss smoothing approach on a variety of standard tasks from D4RL
\citep{fuD4RLDatasetsDeep2020}.
We instantiate loss smoothing on two existing offline-to-online methods. For the MuJoCo
locomotion tasks, the offline phase uses TD3+BC
\citep{fujimotoMinimalistApproachOffline2021} pretraining followed by online
TD3 \citep{fujimotoAddressingFunctionApproximation2018}
fine-tuning. For the AntMaze and Adroit manipulation tasks, we use
SPOT \citep{wuSupportedPolicyOptimization2022a} offline training followed by SPOT online training. We apply smoothing to
both the actor and the critic.
For the TD3/TD3-BC
actor objective, 
since we are smoothing between two different objective functions that can have
drastically different scale, we apply smoothing to both the objective and its scale. 
\[
\mathcal{L}_{\pi,\alpha}^{\mathrm{MuJoCo}}(\theta)
=
(1-\alpha)\mathcal{L}_{\pi}^{\mathrm{TD3+BC}}(\theta)
+
\alpha c_\alpha \mathcal{L}_{\pi}^{\mathrm{TD3}}(\theta),
\]
where $\alpha \in [0,1]$ is the smoothing coefficient. We choose the TD3 actor
loss scale by log-space interpolation,
$
c_\alpha
=
\exp\left((1-\alpha)\log c_0 + \alpha \log 1\right),
$
where $c_0$ matches the scale of the online actor loss to the offline TD3+BC
regime at the start of online fine-tuning, and $c_\alpha=1$ recovers raw,
unconstrained TD3 at the end. This multiplicative annealing is appropriate when
the actor objectives differ by orders of magnitude. For AntMaze and Adroit, loss
smoothing interpolates between the SPOT offline objectives (computed with offline data)
and SPOT online objectives (computed with online data). We provide further details in
\Cref{app:o2o_details}.

We train each agent for 1M offline gradient steps followed by 500k online
environment steps, using five seeds per environment. We compare against the
offline pretrained policy (Offline), online RL from scratch (Online), abrupt
fine-tuning from the pretrained agent (Hard-Switch), and offline-to-online baselines
SPOT \citep{wuSupportedPolicyOptimization2022a}, AWAC
\citep{nairAWACAcceleratingOnline2021}, Cal-QL
\citep{nakamotoCalQLCalibratedOffline2023}, and OPT
\citep{shinOnlinePreTrainingOfflinetoOnline2025}. We standardize
the online interaction budget, evaluation protocol, offline datasets, and
implementation pipeline across methods; stochastic-policy baselines use
stochastic versions of the same TD3-style MLP profiles where the algorithm
requires action log-probabilities.
We report D4RL normalized scores for individual
tasks in \Cref{tab:main-results}, and aggregate performance with rliable
\citep{agarwalDeepReinforcementLearning2021a}
interquartile means with 95\% confidence intervals in \Cref{fig:o2o_rliable}.

\Cref{fig:o2o_rliable} shows that loss smoothing performs well across the
benchmark rather than specializing to a single regime. It achieves the best
overall score, the best suite average on MuJoCo and AntMaze, and remains
competitive on the Adroit tasks. The environment-level trends illustrate the
source-to-target adaptation trade-off in offline-to-online RL. 
On AntMaze Medium-Diverse (\Cref{fig:o2o_sel_curves}, right), online
training from scratch receives zero average score, so the offline initialization
is essential for exploration. On a task like Random HalfCheetah
(\Cref{fig:o2o_sel_curves}, left), however,
conservative baselines that remain close to pretraining inherit limitations of
poor offline data. Loss smoothing provides a middle ground: it preserves the
useful structure of pretraining early in fine-tuning, but gradually removes the
offline constraint as online data becomes available.

\subsection{LLM Fine-Tuning}
\label{sec:llm}
Our next case study considers the overtrained language model setting of
\citet{springerOvertrainedLanguageModels2025b}. Their results show that longer
pretraining can improve the pretrained model while making the resulting
checkpoint harder to adapt, a phenomenon they call \emph{catastrophic overtraining}.
They attribute this degradation to progressive sensitivity: as pretraining
continues, the model's parameters become increasingly sensitive to downstream
updates.
This is a natural fit for our view of adaptation. Instruction tuning abruptly
replaces the broad pretraining distribution with a narrow downstream
distribution, and the effect can be most severe for checkpoints whose parameters
have become highly sensitive to fine-tuning.
$\J_s$ is the pretraining objective and
$\J_t$ is the instruction-tuning objective; moving between them narrows both the
input sequence distribution $u$ and the target-token learning signal $s$.

\begin{figure}
    \centering
    \includegraphics[width=\linewidth]{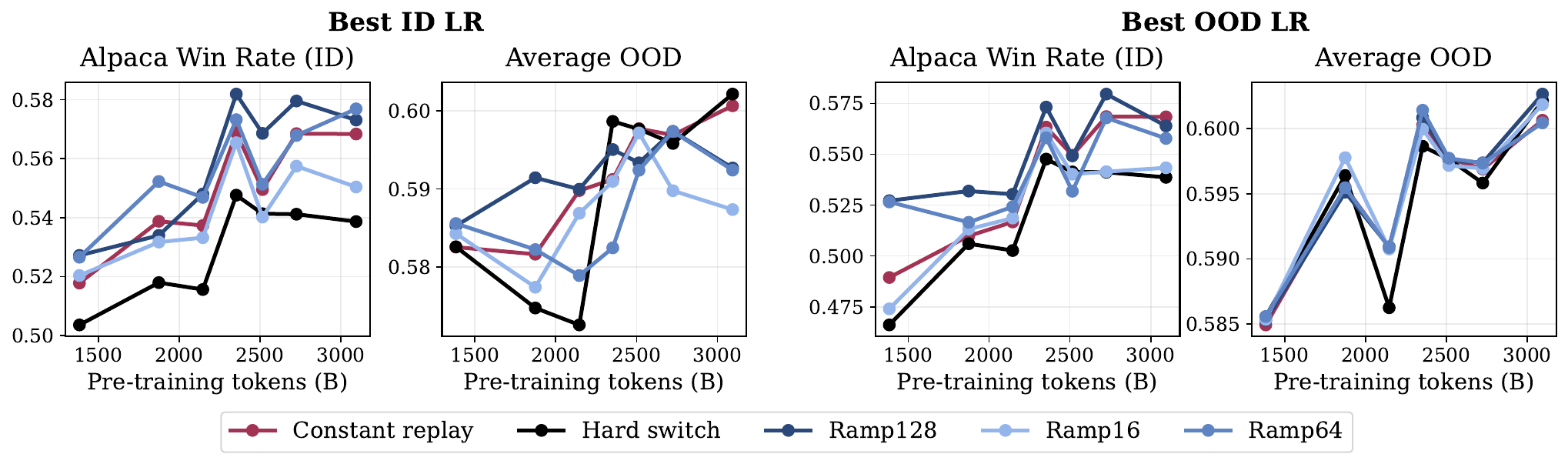}
    \subvspace{-0.7em}
    \includegraphics[width=\linewidth]{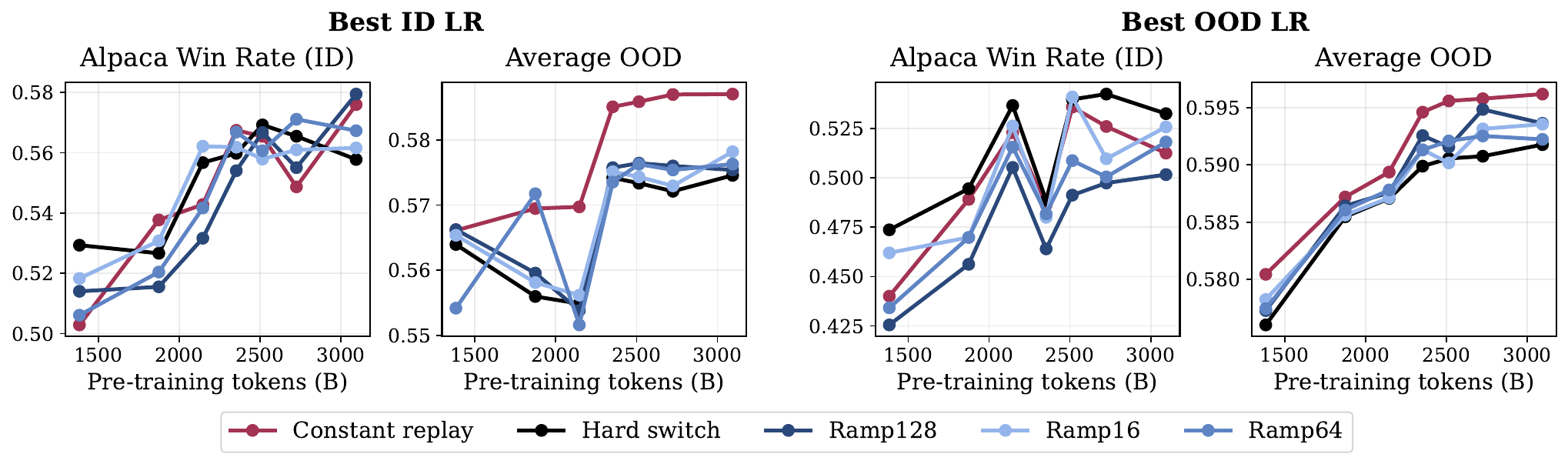}
    \caption{OLMo-1B fine-tuning from checkpoints with different pretraining budgets
    on Anthropic-HH (top) and Tulu-v1 (bottom). Panels compare hard-switch
    fine-tuning, loss smoothing, and constant replay under ID-selected and
    OOD-selected learning rates.}
    \label{fig:llm-main}
\end{figure}

We evaluate whether smoothing this transition reduces the overtraining effect.
We fine-tune OLMo-1B \citep{groeneveldOLMoAcceleratingScience2024} checkpoints
taken at different pretraining budgets, on two
instruction-tuning targets: Anthropic-HH
\citep{baiTrainingHelpfulHarmless2022} and Tulu-v1
\citep{wangHowFarCan2023}. We compare ordinary fine-tuning (Hard-Switch) to
loss smoothing, which gradually anneals from the pretraining loss to the
instruction-tuning loss, and to a constant replay baseline that mixes a fixed
fraction of pretraining examples throughout fine-tuning. Constant replay 
is a useful diagnostic because it is an increasingly commonly used method in LLM
fine-tuning \citep{kothaReplayingPretrainingData2026} that also softens the
distributional change between pretraining and instruction tuning.
For each method, we run three seeds and sweep learning rates in
$\{10^{-5}, 5\cdot10^{-5}, 10^{-4}\}$. See further
details in \Cref{app:llm_details}.

For each target, we report both in-distribution (ID) instruction-following
performance and out-of-distribution (OOD) generalization. ID performance is
measured with AlpacaEval win rate
\citep{alpaca_eval}, while OOD performance is the average
over standard evaluation tasks from the language-model evaluation harness
\citep{eval-harness}, including ARC-Challenge, ARC-Easy,
HellaSwag, PIQA, and WinoGrande. We sweep learning rates and report two
model-selection rules: one that selects the learning rate with the best ID
performance, and one that selects the learning rate with the best OOD
performance. This separates two practical questions: whether smoothing helps
when the practitioner optimizes for the fine-tuning target, and whether it also
helps when preserving general capabilities is the primary goal. 

\Cref{fig:llm-main} shows that loss smoothing improves this trade-off,
especially in the overtrained regime.
When the learning rate is selected for ID performance, loss smoothing and replay
retain strong ID performance while showing less degradation from overtraining
than hard-switch fine-tuning. When the learning rate is selected for OOD
performance, the smoothed variants match or outperform the hard-switch baseline
on OOD metrics. The fact that constant replay also improves the transition supports the
broader idea that methods that soften 
the distributional discontinuity between pretraining and instruction tuning reduce
overtraining-induced brittleness.
Interestingly, while hard-switch performance decreased across pretraining budget,
consistent with what was shown in \citet{springerOvertrainedLanguageModels2025b},
smoothed models maintained or even increased performance, with the single best
performance on Tulu-v1 coming from the final, ostensibly most overtrained checkpoint.
This suggests that catastrophic overtraining is not solely determined by the pretrained
representation or the final instruction-tuning objective, but also by the path used to
move between them.

\subsection{Online Reinforcement Learning}

Finally, we evaluate loss smoothing in online reinforcement learning settings. RL
provides a
testbed for evaluating adaptation methods on more natural and complex distribution
shifts. Not only does the input data distribution change as the agent learns and
explores the environment, but if the agent uses bootstrapping-based methods, then
the output distribution also changes as the agent's value function estimates change over
time. 
Each training phase uses a recent policy-induced distribution over transitions $u$
and value targets $s$, rather than an externally specified task boundary. 
Furthermore, unlike our previous case studies where there were clear boundaries
where the training distribution shifted, in RL the shift boundaries themselves are much
fuzzier. 
\begin{figure}
    \centering
    \subvspace{-2.3ex}
    \begin{subfigure}[c]{.33\linewidth}
        \includegraphics[width=\textwidth]{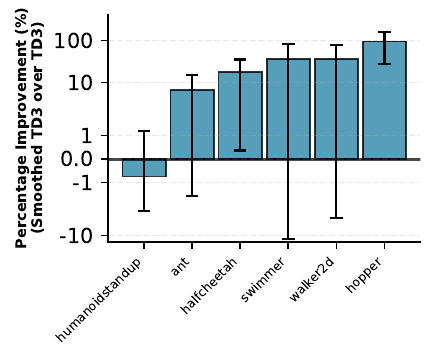}
        \label{fig:td3_results}
        \end{subfigure}
    \begin{subfigure}[c]{.38\linewidth}
        \includegraphics[width=\textwidth]{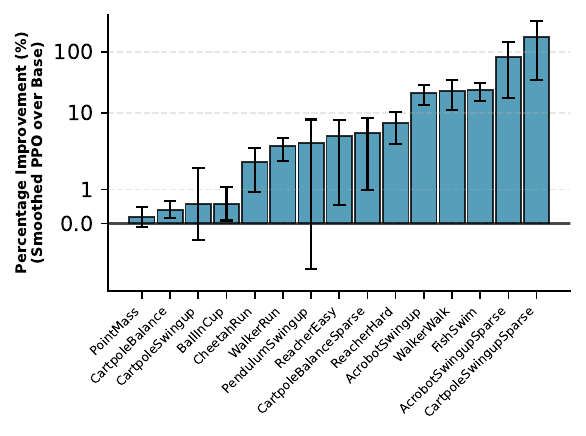}
        \label{fig:ppo_results}
    \end{subfigure}
    \subvspace{-1.2\baselineskip}
    \caption{Percentage improvements of smoothed agents over baseline agents
     for TD3 on Brax environments \textbf{(left)} and PPO on DeepMind Control Suite environments \textbf{(right)}.}
    \label{fig:rl_results}
    \subvspace{-1.2\baselineskip}
\end{figure}

We evaluate the applicability of our method to both off-policy (TD3
\citep{fujimotoAddressingFunctionApproximation2018})
and on-policy (PPO \citep{schulmanProximalPolicyOptimization2017}) algorithms,
testing them on two suites of
continuous control benchmarks: Brax \citep{brax2021github} and 
DeepMind Control Suite (DMC) \citep{tassa2018deepmind} implemented in the Mujoco Playground
\citep{mujoco_playground_2025}, respectively. 
While we could employ loss smoothing on both the policy and
critic losses, we focus on the critic in this section since previous work has shown that
critic adaptation can be more difficult than policy adaptation
\citep{maRevisitingPlasticityVisual2023a}. 
Furthermore, since our algorithm requires knowledge of task shifts that are not
immediately apparent in RL, we simply define a shift as happening every set number of
training steps for each task, denoted by a hyperparameter $n$.

While TD3 already has stabilization mechanisms in the replay
buffer and target network, our experience replay buffer is separate and
stores the previous \emph{outdated} targets directly, rather than recomputing the targets based
on the current target and policy networks, enabling a more direct interpolation
between the old and new objectives. Similarly, with PPO, we save a batch of previous
data and corresponding outdated targets to compute the interpolation. We find that for
both algorithms, despite the already existing mechanisms to promote network stability,
loss smoothing provides significant improvements over the baseline,
with some environments seeing nearly double the performance (\Cref{fig:rl_results}).
For TD3, we find loss smoothing provides improvements on 5 out of 6 Brax environments, 
and for PPO, we find that it provides improvements on 15 of the 18 DMC environments that
we evaluate on (neither agent made significant progress on the 3 Humanoid environments,
excluded from the Figure).
Because these results rely on tuning a boundary that is itself difficult to define in
online RL, we view them as exploratory evidence that loss smoothing can help in a
complex, continuously nonstationary optimization setting. Automatically identifying when
such smoothing boundaries should occur is an important challenge that we leave to future
work.


 


\section{Smoothing Dynamics}
\label{sec:analysis}

The preceding experiments show that loss smoothing can improve adaptation, but they do
not by themselves identify the mechanism. We therefore analyze these dynamics in three
ways. First, we look at which parts of the model move during adaptation, showing that
loss smoothing can preserve task-shared features while concentrating updates in
task-specific components. Second, we measure the \emph{stability gap}: the transient
collapse in previous-task performance immediately after a task boundary. Finally, we
test whether the same retention-adaptation trade-off appears when adapting an
ImageNet-pretrained ViT to DomainNet.

\subsection{Continual Supervised Learning}

\begin{figure*}[t]
    \centering
    \includegraphics[width=0.98\textwidth]{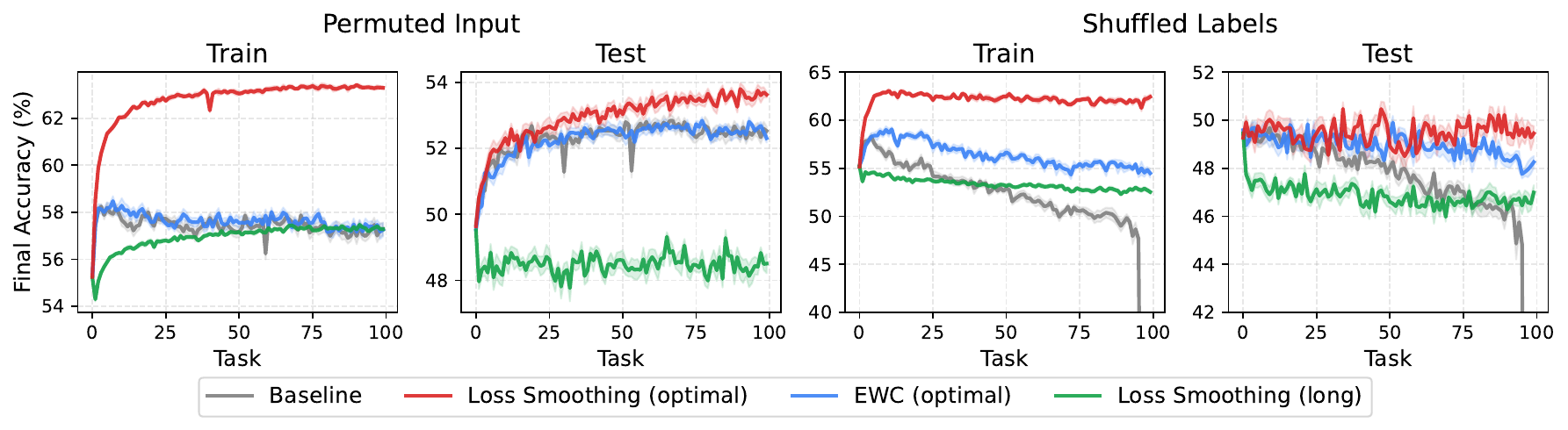}
    \subvspace{-.3em}
    \caption{Supervised settings: resulting accuracy on a sequence of tasks.}
    \label{fig:supervised_main_results}
    \subvspace{-1.1em}
\end{figure*}

We now consider synthetic supervised adaptation setups with controlled data distribution
shifts. We base our experiments on CIFAR-10~\citep{1572824499126417408} and use two
standard transformations~\mbox{\citep{patilExperimentalDesignNonstationary2024a}}:
\emph{permuted input} and \emph{shuffled labels}. 
Although these are simplified settings, they allow us to cleanly isolate the effects of
distribution shifts in just the inputs or outputs, and to easily compare the internal changes
in the model during training.

In the notation of \Cref{sec:method}, adjacent tasks define the source and
target objectives; permuted inputs isolate a change in $u$, while shuffled labels
isolate a change in the learning signal $s$.
In the permuted-input setting, a randomly generated permutation in the input space,
$P_k^{x} : \mathcal{X} \rightarrow \mathcal{X}$, reorders pixel locations. Task $k$ is
defined by the transformed dataset $\mathcal{D}_k = \left\{ \left(P_k^{x}(x_i),
y_i\right) \right\}_{i=1}^n$. 
In the shuffled-label setting, a randomly generated permutation in the output space,
$P_k^{y} : \mathcal{Y} \rightarrow \mathcal{Y}$, remaps the labels, and task $k$ is
defined as $\mathcal{D}_k = \left\{ \left(x_i, P_k^{y}(y_i)\right) \right\}_{i=1}^n$. 
In both settings, after the task switch, the model is only trying to learn the new task,
and not preserve performance on the previous task. Each seed is run on a sequence of 100
independently generated tasks, and we show the average performance with standard deviation.

In Figure~\ref{fig:supervised_main_results}, we compare loss smoothing to a baseline,
both with optimal hyperparameters. Loss smoothing outperforms the baseline on the test
set, with larger gains on later tasks. Interestingly, training accuracy is also higher,
suggesting that the improvement stems from better trainability, despite the additional
regularization to the previous task. 

{\bf Main components} To demonstrate the importance of the two main design choices of
our method—\emph{how} regularization is applied (via loss smoothing) and \emph{when} it
is applied (only at the beginning of training on each task)—we add two additional
baselines in Figure~\ref{fig:supervised_main_results}. 
First, we compare to an EWC~\citep{kirkpatrickOvercomingCatastrophicForgetting2017} variant
of our method, which follows the same training
procedure and regularization coefficient schedule but changes the regularization term from
the loss on the replay data to EWC regularization. In principle, EWC should also
encourage preservation of task-shared weights while allowing task-inconsistent weights
to change. However, it is substantially less effective in our experiments. We
hypothesize that this is due to the local nature of EWC, which makes it ineffective
after the first few steps. 
Second, we compare with a variant of our method with a long decay schedule, where the number of loss
smoothing steps $\tau$ is equal to the whole length of training, making it closer
to standard replay methods from continual learning
\citep{rolnickExperienceReplayContinual2019a}.  
The long decay schedule leads to a significant performance degradation in both setups. We
attribute this to excessive emphasis on old data, which prevents the model from
effectively adapting to the new task. 

\begin{figure*}[t]
    \centering
    \includegraphics[width=0.98\textwidth]{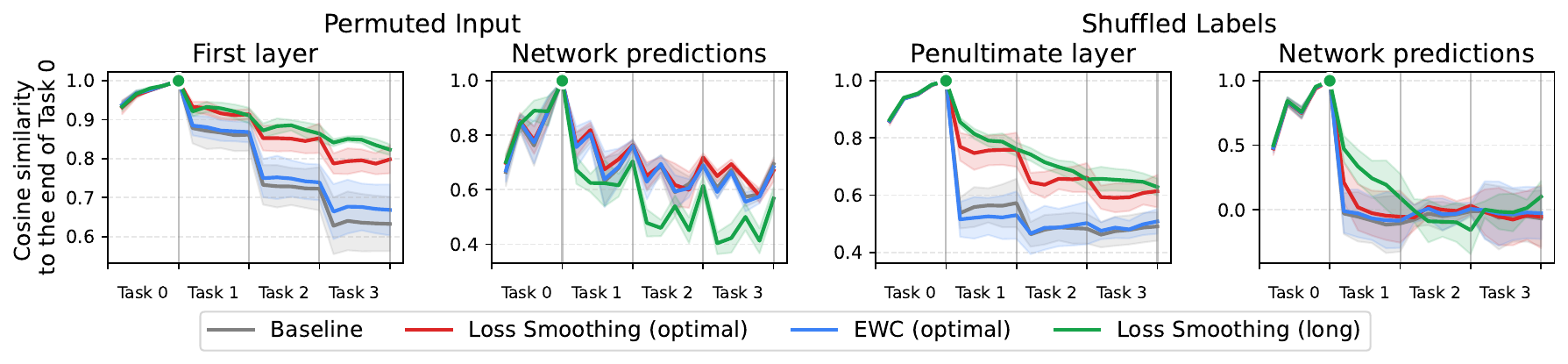}
    \subvspace{-.5em}
    \caption{Supervised settings: analysis of the drift in network activations from the
    final checkpoint of Task~0 (indicated by the round marker) as the model is trained
    on subsequent tasks.}
    \label{fig:supervised_feature_analysis}
\end{figure*}

{\bf Method analysis} To validate our initial motivation that loss smoothing facilitates
adaptation by helping the model identify which components to preserve and which to
modify, in Figure~\ref{fig:supervised_feature_analysis} we analyze the drift in network
activations from the final checkpoint of Task~0 as the model is trained on subsequent
tasks. 
To compare two models, we take a random subset of examples and compute layer-wise
activations for each model via a forward pass. For each layer, we then compute the
average cosine similarity between the corresponding activations. In the permuted-input
setting, we apply the permutation associated with the model checkpoint's task to the inputs
before the forward pass. 

In the permuted-input setting, the functions implemented by all layers except the first
one can, in principle, be shared across tasks, whereas the first layer cannot due to the
different pixel orderings.  
We observe that the first-layer activations change substantially less under loss
smoothing than under the baseline, while the final predictions change to a similar
extent for both methods. This suggests that loss smoothing preserves the functions of
the deeper layers by primarily adapting the first layer to reproduce the appropriate
activations under the new input permutation. 

In the shuffled-label setting, all layers except the last one can be shared across
tasks, since only the image-label correspondence changes.  
Here, the penultimate-layer activations change much less with loss smoothing than with
the baseline, while the final predictions again change similarly for both methods. This
indicates that loss smoothing preserves the shared representation and primarily adapts
the last layer, whereas the baseline modifies the network more uniformly. 

Additionally, we observe that EWC is ineffective at redistributing changes across the
model, while loss smoothing with a long decay schedule alters the final predictions too
strongly and prevents the model from learning an appropriate solution for the current
task. In Appendix~\ref{app:sup_ablations}, we further show that the positive effect of
loss smoothing cannot be explained by implicit learning rate warm-up or increased data
per optimization step, and provide additional ablations of hyperparameter effects.

\subsection{Stability gap}
Beyond final performance, abrupt task switches can also create a
\emph{stability gap}: a transient drop in performance on previous tasks immediately
after the learner begins training on a new task \citep{lange2023continual,harunOvercomingStabilityGap2024}.
This is distinct from final forgetting. A model may eventually recover some previous-task
performance, but the first updates after a distribution shift can still temporarily
destroy useful behavior.

\begin{figure}[t]
    \centering
    \includegraphics[width=.48\textwidth]{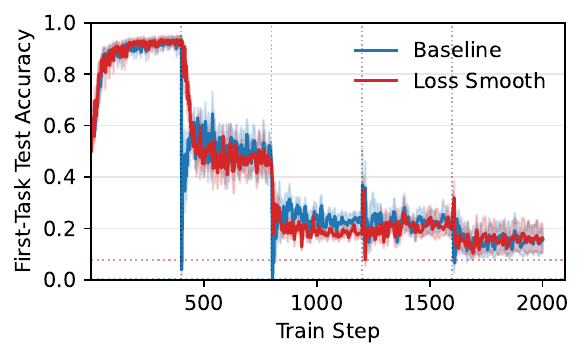}
    \caption{Loss smoothing mitigates the stability gap: the transient drop in
    previous-task performance after training switches to a new task.}
    \label{fig:supervised_stability_gap}
    \subvspace{-1em}
\end{figure}

We evaluate this effect on Split-CIFAR-10, where the ten CIFAR-10 classes are
partitioned into a sequence of five two-class tasks and the model is trained on
the tasks sequentially. This setting creates sharp class-distribution changes at
known task boundaries while preserving a shared visual input space, making it a
direct test of whether the optimizer can incorporate the new classes without
temporarily disrupting features that remain useful for earlier classes. We track
previous-task accuracy throughout training, with particular attention to the
updates immediately after each task switch. As shown in
\Cref{fig:supervised_stability_gap}, loss smoothing substantially reduces the transient
performance drop compared to the hard-switch baseline.

To apply loss smoothing in this setting, we use a per-sample replay weight. Let
$\rho$ denote the target replay weight. When an example from the current task is added
to the replay buffer, we anneal its weight from $1-\rho$ to $\rho$ rather than changing
its contribution abruptly. Conversely, when training begins on a new task, examples from
the new task are annealed from weight $0$ to $1-\rho$. Thus, for an example $z_i$ with
task-boundary time $t_i$, we use a ramp $a_i(t)=\min((t-t_i)/\tau,1)$ and set
\[
    \begin{aligned}
    w_i^{\mathrm{new}}(t) &= a_i(t)(1-\rho),\\
    w_i^{\mathrm{replay}}(t) &= (1-a_i(t))(1-\rho)+a_i(t)\rho.
    \end{aligned}
\]
This per-sample form is the same idea as the global objective interpolation in
\Cref{sec:method}, but applied at the granularity of replay membership. It shows that
loss smoothing is not only a way to improve final adaptation, but also a way to control
the transient optimization path at the moment a distribution shift occurs. 

The reduced stability gap also supports the feature-preservation view of loss
smoothing. If a hard switch forces early updates to rapidly reorganize internal
representations, previous-task accuracy can collapse even before final forgetting is
measured. By reducing this transient collapse, loss smoothing suggests that useful
features are preserved not only at the final checkpoint, but throughout the
optimization trajectory while the model moves from one task objective to the next.
The DomainNet experiment below asks whether this improved path also changes the final
adaptation-retention frontier in a larger pretrained vision model.

\subsection{Vision Transformer DomainNet Adaptation}
\label{sec:domainnet}
\begin{figure*}[t]
    \centering
    \includegraphics[width=0.6\textwidth]{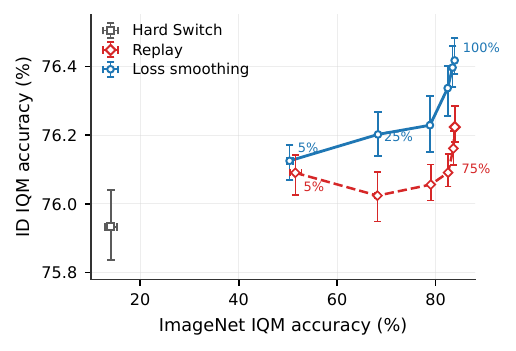}
    \subvspace{-.3em}
    \caption{Single-source DomainNet adaptation from an ImageNet-pretrained ViT.
    We report in-domain DomainNet accuracy against retained ImageNet accuracy.
    Points are IQMs over three seeds and six source domains; error bars are 95\%
    bootstrap confidence intervals. Duration-sweep points correspond to applying
    the source objective for 5, 10, 25, 50, 75, or 100\% of training.}
    \label{fig:domainnet_iqm_tradeoff}
    \subvspace{-1.1em}
\end{figure*}

We next test whether the same preservation-adaptation dynamics persist in a larger
pretrained vision model. Starting from an ImageNet-pretrained ViT, we adapt to each
DomainNet domain separately and measure the trade-off between in-domain DomainNet
accuracy and retention of the original ImageNet classifier. In the notation of
\Cref{sec:method}, ImageNet classification defines the source objective and
single-source DomainNet training defines the target objective; further experimental
details are in \Cref{app:domainnet}. This complements the stability-gap analysis:
there we measured transient retention along the path, while here we compare the final
retention-adaptation frontier induced by different paths.

\Cref{fig:domainnet_iqm_tradeoff} shows that both constant ImageNet replay and loss
smoothing improve over the hard switch, which rapidly loses ImageNet performance. This
supports the central claim that information from the previous objective remains useful
during adaptation: preserving the source task does not merely protect ImageNet
accuracy, but also slightly improves in-domain DomainNet accuracy relative to abrupt
fine-tuning.

The comparison between loss smoothing and matched-duration replay isolates the role of
the path to the target objective. When the ImageNet objective is used for the same
fraction of training, loss smoothing achieves better DomainNet adaptation than replay,
indicating that annealing the source objective away is more effective than keeping a
fixed replay weight and then abruptly removing it. The duration sweep also shows that
the source objective need not remain active for all of training: using it for only 5,
10, or 25\% of training already recovers much of the ImageNet retention lost by the hard
switch, while longer smoothing horizons continue to improve the retention-adaptation
frontier.

\section{Discussion}
\label{sec:discussion}

Across supervised task shifts, pretrained vision adaptation, offline-to-online RL,
online RL, and LLM fine-tuning, we find that abrupt changes in the input distribution
$u$, learning signal $s$, or loss family $\ell$ can make 
the updates after the shift brittle. Loss smoothing reduces this brittleness by keeping
early updates partially tied to the source objective while the model begins to incorporate
target-task information. 

The case studies in our work show that the source objective often contains
useful structure. Supervised replay preserves reusable features,
offline RL pretraining can provide behaviors and value estimates that help early
online learning, and 
language-model pretraining supports broad capabilities that
can be damaged by narrow instruction tuning. Loss smoothing keeps these sources of
structure active at the start of adaptation, then anneals them away so that the model
can specialize.

\paragraph{Limitations} The method requires access to, or an estimate of,
the source objective, which may involve replay data, stored targets, or continued
access to previous training data. It also introduces schedule choices, and the
right smoothing boundary is especially unclear in online RL, where the data
distribution changes continuously with the policy. When loss families have very
different units, as in TD3+BC-to-TD3 fine-tuning, their scales must also be
handled carefully. Compared to normal training, loss smoothing also requires more
compute at the start of adaptation.
Finally, our ablations show that smoothing for too long can
hurt by keeping the learner tied to outdated data. Automatically detecting when
to smooth, how long to smooth, and how to normalize incompatible objectives are
important directions for future work.

\section*{Acknowledgments}
We would like to thank Andrei Mircea for the
valuable comments and discussions.
We also thank Mila (\url{mila.quebec}) and its IDT team for providing and supporting the
computing resources used in this work. Ekaterina Lobacheva is supported by IVADO and the
Canada First Research Excellence Fund. Sarath Chandar is supported by the Canada CIFAR
AI Chairs program, the Canada Research Chair in Lifelong Machine Learning, and the NSERC
Discovery Grant.  

\bibliographystyle{plainnat}
\bibliography{ref}
\clearpage
\appendix

\section{Continual Supervised Learning}
\subsection{Experimental Details}
\label{app:sup_setup}
We use a simple 3-layer MLP: input layer (3072 → 100) → ReLU → hidden layer (100 → 100) → ReLU → output layer (100 → 10). The model is trained with SGD, batch size 256, for 20k steps per task. We run all experiments with 10 different random seeds and show average results with standard deviation in all the plots.

We ran baseline, loss smoothing, and EWC experiments across several learning rates and weight decay values, and found that a learning rate of 0.01 and weight decay of 0.001 were optimal for all methods. These values are used in all experiments.  
For loss smoothing, we additionally search for the optimal number of smoothing steps from $[100, 500, 1000, 5000, 10000, 20000]$, finding it to be 500 for permuted input and 5000 for shuffled labels. A similar search for EWC yields optimal regularization steps of 500 for permuted input and 100 for shuffled labels.

\subsection{Ablations}
\label{app:sup_ablations}

\begin{figure*}[t]
    \centering
    \includegraphics[width=0.98\textwidth]{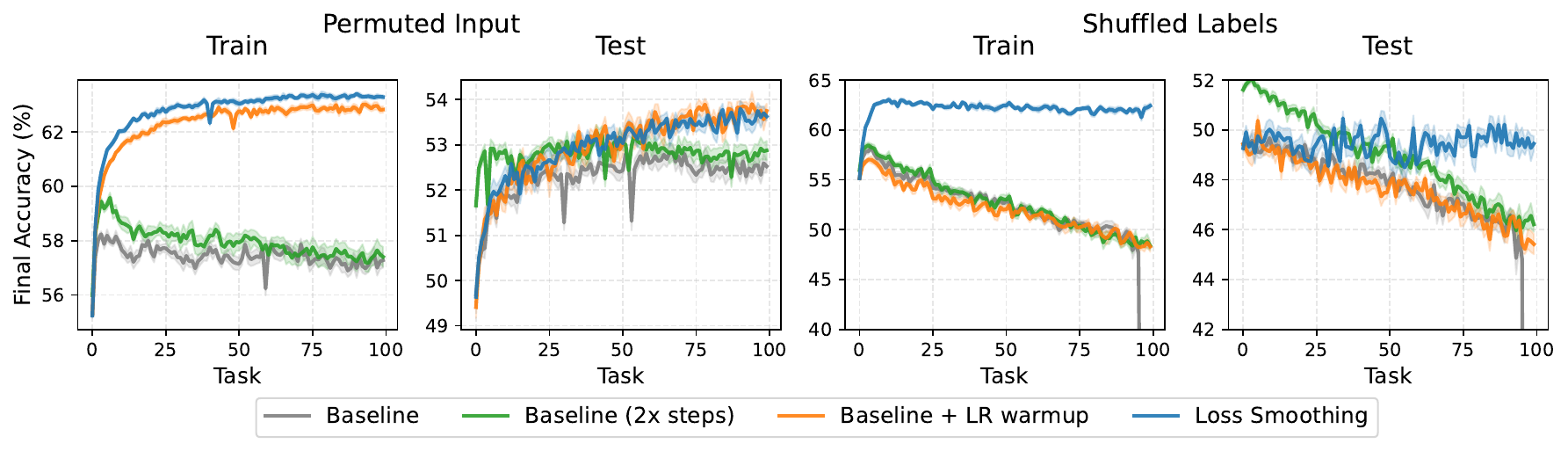}
    \caption{Ablation in supervised settings: comparison to baseline variants using learning rate warm-up or increased data per optimization step.}
    \label{fig:sup_ablation_alternatives}
\end{figure*}

The positive effect of loss smoothing could potentially arise from using a learning rate warm-up at the start of each task or from having more data per optimization step. In Figure~\ref{fig:sup_ablation_alternatives}, we compare to the baseline with these modifications. For the baseline with learning rate warm-up, we use the same schedule for the current-batch loss coefficient as in loss smoothing, but set the replay term to zero. Warm-up helps in the permuted-input setting, achieving results comparable to loss smoothing, but has little effect for shuffled labels. For the baseline with more data, we double the batch size to match the data size used per optimization step in loss smoothing. While this slightly improves optimization, it does not replicate the benefits of loss smoothing.

\begin{figure*}[t]
    \centering
    \includegraphics[width=0.98\textwidth]{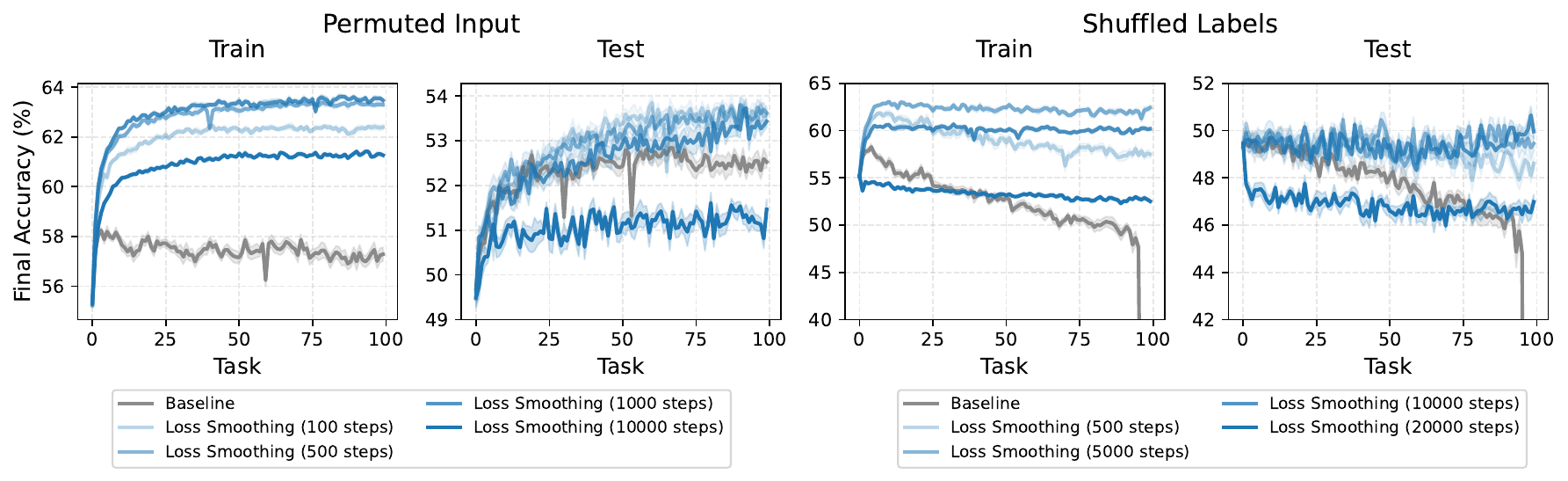}
    \caption{Ablation in supervised settings: effect of the number of loss smoothing steps on model performance.}
    \label{fig:sup_ablation_decay}
\end{figure*}

\begin{figure*}[t]
    \centering
    \includegraphics[width=0.98\textwidth]{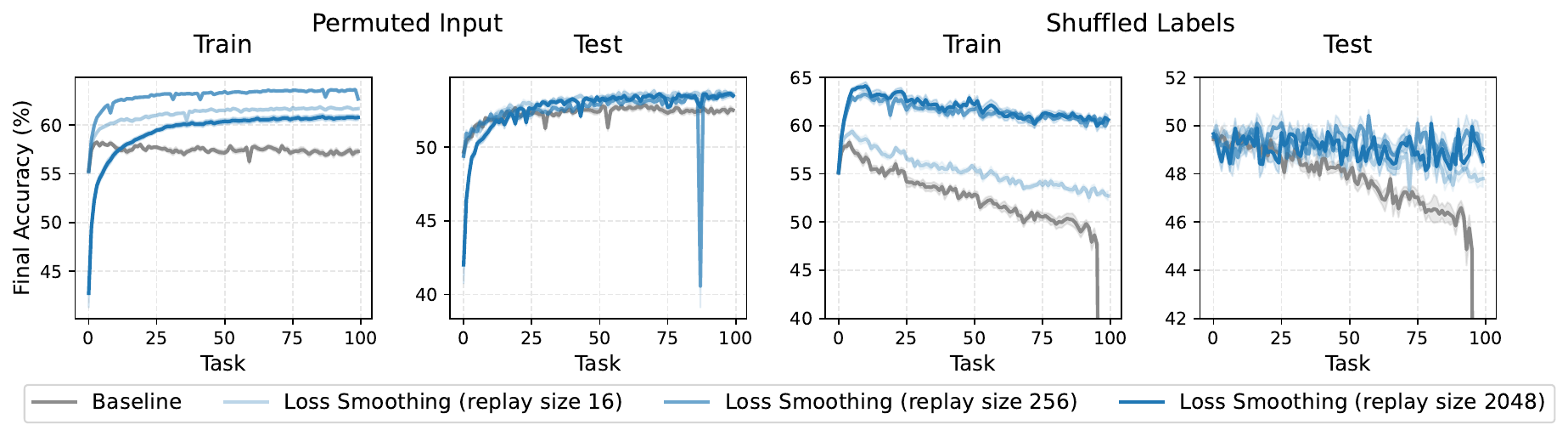}
    \caption{Ablation in supervised settings: effect of the amount of replay data on model performance.}
    \label{fig:sup_ablation_bs}
\end{figure*}
In Figures~\ref{fig:sup_ablation_decay} and~\ref{fig:sup_ablation_bs}, we analyze the
effects of two main hyperparameters of loss smoothing: the number of smoothing steps
$\tau$ and the amount of replay data.
For the number of steps, there exists an optimal value that depends on the task (here,
500 for permuted input and 5000 for shuffled labels). Fewer steps reduce the positive
effect of loss smoothing, while too many steps degrade performance by overemphasizing
older data. 
For the amount of replay data, a batch of 256 examples is generally sufficient. Using
fewer examples can hurt performance, while using more does not provide additional
benefits. 

\section{DomainNet Experimental Details}
\label{app:domainnet}

We run the full-DomainNet ViT experiments with a \texttt{vit\_base\_patch16\_224} model
initialized from ImageNet pretraining. We keep the pretrained 1000-way ImageNet head,
remove the classifier from the shared backbone, and add a randomly initialized 345-way
DomainNet head. Thus, adaptation changes the shared representation and DomainNet head,
while ImageNet retention is measured with the preserved ImageNet head.

We use all six DomainNet domains: clipart, infograph, painting, quickdraw, real, and
sketch. For the single-source runs in \Cref{fig:domainnet_iqm_tradeoff}, each run trains
on one source domain and evaluates in-domain accuracy on that same domain's official
test split. We construct a stratified 10\% validation split from the official training
split for model selection and leave the official test split unchanged. All images are
resized and cropped to $224\times224$ using ImageNet normalization.

All methods are trained for 60 epochs with batch size 256, AdamW, learning rate
$2\times10^{-5}$, weight decay $0.1$, 5\% linear learning-rate warmup, and cosine decay.
The hard-switch baseline minimizes only the DomainNet cross-entropy loss after
adaptation begins. Replay and loss smoothing additionally use ImageNet training batches
through the original ImageNet head. Let $\lambda_t$ be the ImageNet loss coefficient at
step $t$; the optimized loss is
\[
    (1-\lambda_t)\mathcal{L}_{\mathrm{DomainNet}}
    + \lambda_t\mathcal{L}_{\mathrm{ImageNet}} .
\]
Replay uses a constant $\lambda_t=0.5$ throughout training. Loss smoothing starts with
$\lambda_0=1$ and linearly anneals to $0$ over a specified fraction of training: 5, 10,
25, 50, 75, or 100\%. We also run matched-duration replay controls, where
$\lambda_t=0.5$ is held constant for the same initial fraction of training and then set
to $0$.

We summarize final epoch-60 checkpoints. For each method, we form a $3\times6$ score
matrix from three seeds and six source domains. We report the interquartile mean (IQM)
with 95\% percentile bootstrap confidence intervals using 50{,}000 bootstrap samples.
The bootstrap resamples seeds while keeping the six source domains
as the task strata.

\begin{table}[t]
    \centering
    \caption{DomainNet single-source ViT results used in
    \Cref{fig:domainnet_iqm_tradeoff}. Values are IQM accuracies with 95\% bootstrap
    confidence intervals.}
    \label{tab:domainnet_iqm}
    \begin{tabular}{lcc}
        \toprule
        Method & DomainNet ID & ImageNet retention \\
        \midrule
        Hard switch & 75.93 [75.84, 76.04] & 14.01 [12.77, 15.35] \\
        Replay, 100\% & 76.22 [76.18, 76.29] & 83.90 [83.78, 84.00] \\
        Replay, 5\% & 76.09 [76.03, 76.14] & 51.50 [50.41, 52.75] \\
        Replay, 10\% & 76.02 [75.95, 76.09] & 68.18 [67.57, 68.62] \\
        Replay, 25\% & 76.06 [76.01, 76.11] & 79.01 [78.81, 79.20] \\
        Replay, 50\% & 76.09 [76.05, 76.15] & 82.53 [82.40, 82.66] \\
        Replay, 75\% & 76.16 [76.11, 76.21] & 83.60 [83.50, 83.70] \\
        Loss smoothing, 5\% & 76.13 [76.07, 76.17] & 50.33 [49.74, 51.02] \\
        Loss smoothing, 10\% & 76.20 [76.14, 76.27] & 68.28 [67.80, 68.57] \\
        Loss smoothing, 25\% & 76.23 [76.15, 76.32] & 78.84 [78.74, 78.94] \\
        Loss smoothing, 50\% & 76.34 [76.26, 76.40] & 82.45 [82.29, 82.61] \\
        Loss smoothing, 75\% & 76.40 [76.34, 76.46] & 83.43 [83.28, 83.58] \\
        Loss smoothing, 100\% & 76.42 [76.38, 76.48] & 83.86 [83.75, 83.98] \\
        \bottomrule
    \end{tabular}
\end{table}


\section{Offline-to-Online RL}
\subsection{Experimental Details}
\label{app:o2o_details}
For the offline-to-online RL case study, offline-initialized methods use 1M
offline gradient steps followed by 500k online environment steps, with five
random seeds per environment.  Our MuJoCo loss-smoothing instantiation uses a
TD3+BC-style offline objective \citep{fujimotoMinimalistApproachOffline2021}
and TD3 online fine-tuning \citep{fujimotoAddressingFunctionApproximation2018}.
Our AntMaze and Adroit loss-smoothing instantiation uses SPOT-style offline
training and SPOT online fine-tuning \citep{wuSupportedPolicyOptimization2022a}.

\paragraph{Tasks and budgets.}
We use the D4RL benchmark suite \citep{fuD4RLDatasetsDeep2020} with the
following dataset identifiers:
\texttt{halfcheetah-medium-v2}, \texttt{hopper-medium-v2},
\texttt{walker2d-medium-v2}, \texttt{halfcheetah-random-v2},
\texttt{hopper-random-v2}, and \texttt{walker2d-random-v2} for MuJoCo
locomotion; \texttt{antmaze-umaze-v2}, \texttt{antmaze-umaze-diverse-v2},
\texttt{antmaze-medium-play-v2}, and \texttt{antmaze-medium-diverse-v2} for
AntMaze; and \texttt{pen-cloned-v1}, \texttt{hammer-cloned-v1},
\texttt{door-cloned-v1}, and \texttt{relocate-cloned-v1} for Adroit.  Unless
otherwise noted, results use seeds $0,\ldots,4$.  Offline-initialized methods use
1M offline gradient updates followed by 500k online environment steps.  The
online-from-scratch control uses the same 500k online-step budget with no offline
pretraining, and the offline control evaluates the pretrained checkpoint without
online updates.

\paragraph{Replay and online updates.}
All online training runs use a single training environment and one gradient
update per environment step.  Evaluation is vectorized, but training is not.
Mini-batches have size 256.  For smoothed TD3+BC-to-TD3 runs, each update draws
one batch from the online replay buffer and one independent batch from the fixed
D4RL dataset; the offline actor/critic losses are computed on the offline batch,
the online actor/critic losses are computed on the online batch, and the losses
are interpolated after they are computed.  The hard-switch and online TD3
controls do not use the offline loss branch.  TD3-family online replay buffers
have capacity 2M transitions.  For SPOT-family support tasks,
non-smoothed SPOT fine-tuning follows the SPOT replay convention by initializing
a mixed replay buffer with offline transitions and then appending online data.
Smoothed SPOT uses a separate online-only replay buffer for the online branch and
samples the offline branch directly from the D4RL data; the mixed and online
SPOT buffers each have capacity 2M.  AWAC
\citep{nairAWACAcceleratingOnline2021} uses a uniform replay buffer initialized
with the offline data and appended online data.
Cal-QL \citep{nakamotoCalQLCalibratedOffline2023} uses separate 2M-transition
offline and online replay buffers and, after online data is available, samples
half of each update batch from each buffer.  OPT
\citep{shinOnlinePreTrainingOfflinetoOnline2025} uses uniform replay in our
implementation: the first 25k online steps are collected with the frozen offline
policy, then the online critic is adapted for 50k critic-only updates, and then
fine-tuning proceeds within the same 500k online interaction budget.  We do not
use the prioritized replay variant of OPT.

\paragraph{Evaluation and aggregation.}
We evaluate every 10k online environment steps, for 50 evaluations over the
500k-step budget.  MuJoCo evaluation uses 128 episodes; AntMaze and Adroit use
10 episodes.  Evaluation returns are computed with the unmodified environment
rewards and are converted to D4RL normalized scores using the standard
\texttt{get\_normalized\_score} convention.  Final-return tables average the
last three evaluation points within each seed and report mean $\pm$ standard
error across seeds for each environment.  Suite-average rows are arithmetic
averages over the environment-level means in the suite, with standard errors
computed across environments.  Learning-curve bands are standard errors across
seeds.  The suite-level IQM plots use rliable
\citep{agarwalDeepReinforcementLearning2021a} with 95\% bootstrap confidence
intervals.

\paragraph{Architectures and optimization.}
Within each algorithm family, all variants share the same network and optimizer
profile.  Dense MuJoCo TD3+BC, TD3, and TD3+OPT runs use two-layer 256-unit ReLU
actor and critic MLPs with layer normalization and deterministic tanh-scaled
actions.  AntMaze SPOT runs use the same two-layer ReLU MLP without layer
normalization.  Adroit SPOT runs use the two-layer ReLU MLP with post-activation
layer normalization and OPT-style small uniform final-layer initialization.
AWAC and Cal-QL use stochastic versions of these TD3-style profiles: the actor
head is replaced with a tanh-squashed Gaussian distribution head so that the
algorithms can evaluate action log-probabilities, while the actor and critic
torsos keep the same two-layer 256-unit ReLU architecture and domain-specific
layer-normalization choices as the corresponding TD3/SPOT runs.  The main
optimization hyperparameters are summarized in \Cref{tab:o2o_hyperparameters}.

\begin{table*}[t]
\centering
\caption{Key offline-to-online RL hyperparameters.  Values are shared across
environments unless a domain-specific value is listed.}
\label{tab:o2o_hyperparameters}
\small
\setlength{\tabcolsep}{4pt}
\renewcommand{\arraystretch}{1.08}
\begin{tabular}{@{}p{0.20\textwidth}p{0.74\textwidth}@{}}
\toprule
Component & Values \\
\midrule
TD3-family online training
& Actor learning rate $10^{-4}$; critic learning rate $3\cdot10^{-4}$;
discount $\gamma=0.99$; target-update rate $\tau=0.005$; exploration noise
$0.1$; target policy noise $0.2$; target-noise clip $0.5$; delayed
actor/target updates every two critic updates. \\
TD3+BC offline pretraining
& 1M gradient updates; \texttt{bc\_alpha}=2.5. \\
SPOT support model and critic/actor updates
& Actor learning rate $10^{-4}$; critic learning rate $3\cdot10^{-4}$; the
same TD3 target and exploration parameters; VAE learning rate $10^{-3}$; VAE
hidden dimension 750; 100k VAE updates; $\beta=0.5$; one VAE sample for the
support cost. \\
SPOT support coefficient
& $\lambda_t=\lambda_0\max(\lambda_{\mathrm{floor}},1-t/10^6)$, with
$(\lambda_0,\lambda_{\mathrm{floor}})=(0.05,0.2)$ for AntMaze and
$(1.0,0.5)$ for Adroit.  The SPOT baseline uses this cooling schedule; the
hard-switch support baseline disables it. \\
AWAC
& Actor and critic learning rates $3\cdot10^{-4}$; advantage temperature
$\lambda=0.1$ on AntMaze and $1.0$ otherwise; exponentiated advantages clipped
at 100. \\
Cal-QL
& Actor learning rate $10^{-4}$; critic and entropy-temperature learning rates
$3\cdot10^{-4}$; CQL coefficient 5; 10 sampled actions; CQL temperature 1;
offline calibration.  On AntMaze we use the reference sparse-reward settings:
training reward transform $10r-5$, Lagrange CQL with target action gap 0.8, max
target backup, and CQL-difference lower clipping at $-200$. \\
OPT
& Uniform replay; 25k online collection steps with the frozen offline policy;
50k critic-only online-critic adaptation updates; OPT inner critic learning
rate $3\cdot10^{-4}$.  The mixed-Q actor schedule uses $\kappa=1$ for random
MuJoCo, $0.3\to0.9$ over 150k steps for medium MuJoCo, $0.1\to0.9$ over 100k
steps for AntMaze, and $0.1\to0.9$ over 250k steps for Adroit. \\
\bottomrule
\end{tabular}
\end{table*}

\paragraph{Reward transforms.}
Reward transforms are applied only for training, taken from reference implementations;
evaluation always uses raw
environment rewards before D4RL normalization.  SPOT-family AntMaze runs,
including loss smoothing and SPOT+OPT, use the SPOT reward transform $r'=r-1$.
AWAC uses $r'=r-1$ on AntMaze and raw rewards elsewhere.  Cal-QL uses $r'=10r-5$
on AntMaze and raw rewards elsewhere.  The raw TD3 online-from-scratch control
on AntMaze also uses $r'=10r-5$.

\paragraph{Loss-smoothing schedule.}
Loss smoothing uses an environment-step clock.  Let
$h$ be the smoothing horizon and $\alpha_t=\min(t/h,1)$.  The offline weight is
$1-\alpha_t$, the online weight is $\alpha_t$, and both actor and critic losses
are smoothed:
\[
\mathcal{L}_t = (1-\alpha_t)\mathcal{L}_{\mathrm{offline}}
              + \alpha_t \mathcal{L}_{\mathrm{online}} .
\]
For the main results, dense MuJoCo and Adroit cloned tasks use a 50k-step ramp,
while AntMaze uses a 250k-step ramp.  We also run ramp-length ablations at 10k,
50k, 100k, and 250k steps, with the main-result plots selecting one ramp per
domain.
For MuJoCo TD3+BC-to-TD3 actor smoothing, the online TD3 actor branch is
normalized by a stop-gradient $\mathbb{E}[|Q|]$ scale and multiplied by a
log-space scale annealing factor,
\[
c_t = \exp\left((1-\alpha_t)\log c_0 + \alpha_t\log c_1\right),
\]
where $c_0$ is the offline TD3+BC behavior-cloning coefficient read from the
checkpoint and $c_1$ is the current online TD3 $|Q|$ scale.  This preserves a
smooth handoff between the TD3+BC actor-loss units and the online TD3 actor-loss
units.  Equivalently, if the online actor loss is written in raw TD3 units, this
endpoint corresponds to scale one.  The SPOT-to-TD3 ablation uses analogous
log-space scale annealing when the online branch is changed from the SPOT actor
objective to the unconstrained TD3 actor objective.

\subsection{Additional Results and Ablations}

    \begin{table*}[t]
\centering
\caption{
Normalized scores on individual environments. Values are mean $\pm$ standard error.
Best result in each environment is bolded and second-best is underlined.
Overall averages are computed over the three suite averages.
}
\label{tab:main-results}
\small
\setlength{\tabcolsep}{3.2pt}
\renewcommand{\arraystretch}{1.04}
\newcommand{\s}[2]{#1{\scriptsize$\pm$#2}}
\newcommand{\dash}{---}
\begin{tabular}{@{}lccccccc@{}}
\toprule
Env. & Offline & Online & Hard-Switch & AWAC & Cal-QL & OPT & \shortstack{Loss\\Smoothing} \\
\midrule

\multicolumn{8}{@{}l}{\textit{MuJoCo}} \\
HalfCheetah-M
& \s{47.3}{0.0} & \s{29.5}{7.5} & \s{48.2}{0.3} & \second{\s{52.2}{0.3}} & \best{\s{54.5}{0.6}} & \s{48.5}{0.2} & \s{48.2}{0.5} \\
Hopper-M
& \s{47.2}{1.9} & \s{69.6}{18.6} & \second{\s{101.2}{5.8}} & \s{81.9}{4.1} & \s{74.4}{5.6} & \best{\s{105.5}{0.4}} & \s{99.9}{5.5} \\
Walker2d-M
& \s{83.2}{0.4} & \s{61.5}{13.9} & \second{\s{99.2}{1.9}} & \s{86.4}{0.2} & \s{87.4}{0.4} & \s{94.6}{1.1} & \best{\s{103.1}{0.7}} \\
HalfCheetah-R
& \s{8.3}{0.3} & \s{18.4}{0.6} & \second{\s{43.0}{2.6}} & \s{32.9}{0.6} & \s{13.1}{1.8} & \s{37.5}{1.0} & \best{\s{45.2}{1.7}} \\
Hopper-R
& \s{13.5}{7.1} & \second{\s{67.2}{17.5}} & \s{55.0}{9.8} & \s{18.1}{1.3} & \s{11.9}{2.0} & \s{14.9}{2.1} & \best{\s{77.9}{16.7}} \\
Walker2d-R
& \s{7.2}{3.4} & \second{\s{73.3}{7.5}} & \best{\s{74.7}{3.3}} & \s{4.2}{0.3} & \s{11.0}{3.8} & \s{9.0}{1.7} & \s{68.0}{13.1} \\
\addlinespace[1pt]
\textit{Average}
& \s{34.5}{12.3} & \s{53.2}{9.5} & \second{\s{70.2}{10.4}} & \s{45.9}{13.7} & \s{42.0}{14.1} & \s{51.7}{16.5} & \best{\s{73.7}{10.1}} \\

\midrule
\multicolumn{8}{@{}l}{\textit{AntMaze}} \\
UMaze
& \s{26.0}{16.0} & \s{0.0}{0.0} & \s{40.7}{14.2} & \s{6.7}{1.5} & \best{\s{98.7}{1.3}} & \s{44.0}{21.9} & \second{\s{80.0}{20.0}} \\
UMaze-Div
& \s{18.0}{9.7} & \s{0.0}{0.0} & \s{67.7}{10.6} & \s{14.7}{6.2} & \s{47.3}{3.9} & \best{\s{81.3}{7.2}} & \second{\s{76.0}{19.1}} \\
Medium-Play
& \s{58.0}{7.3} & \s{0.0}{0.0} & \s{55.7}{7.4} & \s{0.0}{0.0} & \s{0.0}{0.0} & \best{\s{84.0}{4.5}} & \second{\s{83.3}{8.0}} \\
Medium-Div
& \s{38.0}{11.1} & \s{0.0}{0.0} & \s{21.7}{7.1} & \s{0.0}{0.0} & \s{28.0}{16.7} & \second{\s{68.0}{10.1}} & \best{\s{94.0}{3.1}} \\
\addlinespace[1pt]
\textit{Average}
& \s{35.0}{8.7} & \s{0.0}{0.0} & \s{46.4}{9.9} & \s{5.3}{3.5} & \s{43.5}{20.8} & \second{\s{69.3}{9.1}} & \best{\s{83.3}{3.9}} \\

\midrule
\multicolumn{8}{@{}l}{\textit{Adroit}} \\
Pen
& \s{47.8}{9.0} & \s{101.6}{8.8} & \s{114.9}{4.0} & \s{55.3}{4.7} & \s{103.0}{7.1} & \best{\s{121.8}{4.7}} & \second{\s{118.2}{3.5}} \\
Hammer
& \s{2.2}{0.4} & \s{100.5}{11.6} & \s{117.3}{1.4} & \s{0.3}{0.0} & \second{\s{120.9}{3.4}} & \s{112.9}{5.8} & \best{\s{122.1}{2.5}} \\
Door
& \s{0.3}{0.2} & \s{58.3}{14.2} & \s{0.3}{0.2} & \s{-0.1}{0.0} & \s{19.9}{14.6} & \best{\s{72.1}{3.8}} & \second{\s{63.4}{10.8}} \\
Relocate
& \second{\s{-0.0}{0.0}} & \s{-0.1}{0.0} & \s{-0.1}{0.0} & \s{-0.2}{0.0} & \s{-0.2}{0.0} & \best{\s{0.3}{0.1}} & \s{-0.1}{0.0} \\
\addlinespace[1pt]
\textit{Average}
& \s{12.6}{11.8} & \s{65.1}{23.9} & \s{58.1}{33.5} & \s{13.8}{13.8} & \s{60.9}{30.0} & \best{\s{76.8}{27.7}} & \second{\s{75.9}{28.7}} \\

\midrule
\textit{Overall}
& 27.3 & 39.4 & 58.2 & 21.7 & 48.8 & \second{65.9} & \best{77.6} \\

\bottomrule
\end{tabular}
\end{table*}

\begin{figure*}[t]
    \centering
    \begin{subfigure}{\textwidth}
        \includegraphics[width=\textwidth]{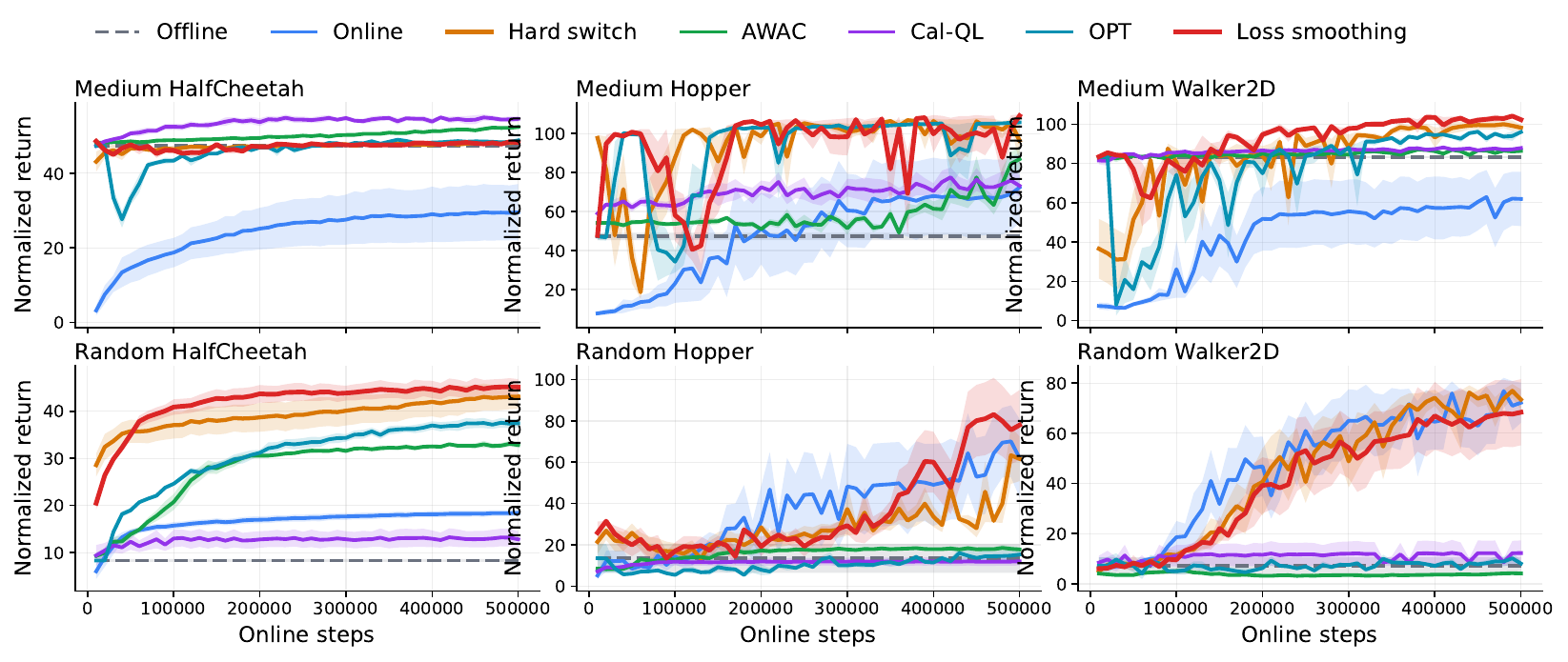}
        \caption{Dense MuJoCo locomotion tasks.}
        \label{fig:o2o_dense_curves}
    \end{subfigure}
    \hfill
    \begin{subfigure}{\textwidth}
        \includegraphics[width=\textwidth]{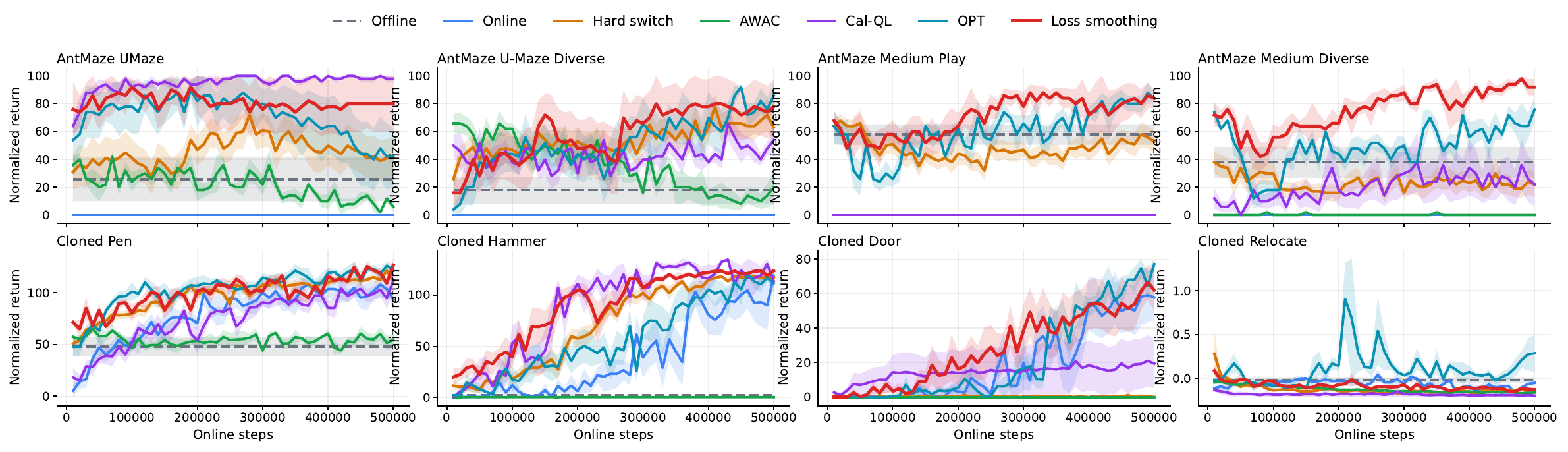}
        \caption{AntMaze and Adroit tasks.}
        \label{fig:o2o_support_curves}
    \end{subfigure}
    \caption{Additional offline-to-online RL learning curves. Shaded regions denote
    standard error across seeds. These curves complement the selected examples in
    \Cref{fig:o2o_sel_curves} by showing the broader suite-level behavior across
    dense locomotion and support-constrained environments.}
    \label{fig:o2o_appendix_curves}
\end{figure*}

We provide additional learning curves in \Cref{fig:o2o_appendix_curves}. In dense MuJoCo
tasks, loss smoothing
typically retains the fast improvement of offline initialization while allowing
the policy to move beyond the constraints of the offline objective. In the
support-constrained AntMaze and Adroit tasks, the curves show the complementary
role of the offline objective: early smoothing stabilizes fine-tuning in regimes
where online data alone can be difficult to use.

\begin{figure*}[t]
    \centering
    \begin{subfigure}{0.32\textwidth}
        \includegraphics[width=\textwidth]{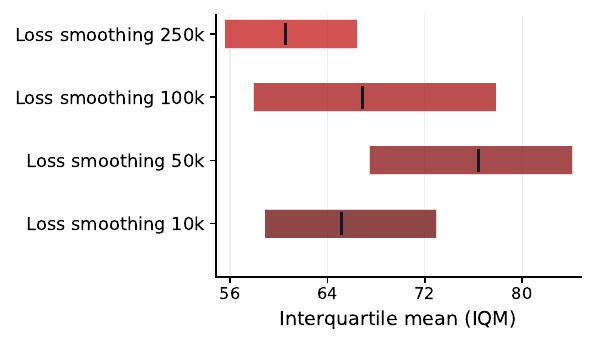}
        \caption{MuJoCo.}
        \label{fig:o2o_ablation_mujoco}
    \end{subfigure}
    \hfill
    \begin{subfigure}{0.32\textwidth}
        \includegraphics[width=\textwidth]{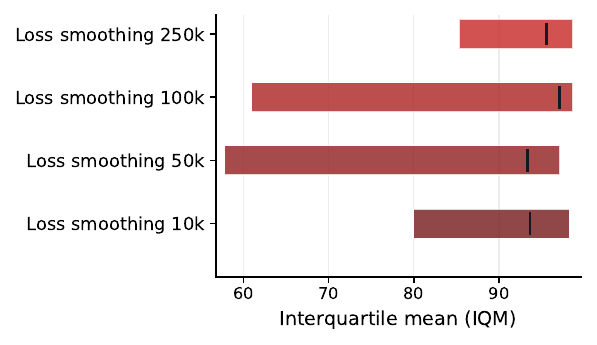}
        \caption{AntMaze.}
        \label{fig:o2o_ablation_antmaze}
    \end{subfigure}
    \hfill
    \begin{subfigure}{0.32\textwidth}
        \includegraphics[width=\textwidth]{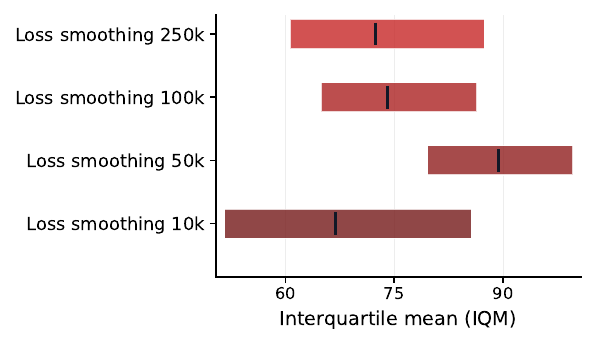}
        \caption{Adroit.}
        \label{fig:o2o_ablation_adroit}
    \end{subfigure}
    \caption{Offline-to-online RL ablations for the loss-smoothing horizon. Each
    panel reports the suite-level interquartile mean with 95\% bootstrap confidence
    intervals. The main results use a domain-specific horizon selected from these
    ablations: 50k environment steps for MuJoCo and Adroit, and 250k environment
    steps for AntMaze.}
    \label{fig:o2o_ablation_iqm}
\end{figure*}

We also ablate the number of environment steps over which the source objective
is annealed away. \Cref{fig:o2o_ablation_iqm} shows that the preferred horizon
depends on the domain. MuJoCo and Adroit benefit from a shorter handoff, where
online interaction quickly provides useful target-task information. AntMaze
benefits from a longer handoff, consistent with the sparse-reward setting where
the offline policy remains important for exploration over a larger fraction of
fine-tuning. These results support treating the smoothing horizon as a domain
hyperparameter rather than a fixed constant.

\section{Overtrained LLM Fine-tuning}
\label{app:llm_details}
\paragraph{OLMo-1B checkpoints.}
We follow the OLMo-1B setting from \citet{springerOvertrainedLanguageModels2025b}.
All runs initialize from public \texttt{allenai/OLMo-1B-hf} checkpoints
\citep{groeneveldOLMoAcceleratingScience2024} and use the corresponding OLMo
tokenizer. We fine-tune all model parameters; no layers are frozen and no
adapter modules are introduced. The checkpoint grid is shown in
\Cref{tab:olmo1b-checkpoints}. We use the token counts in the checkpoint names
as the pretraining budget.

\begin{table}[t]
\centering
\caption{OLMo-1B checkpoints used in the fine-tuning experiments.}
\label{tab:olmo1b-checkpoints}
\begin{tabular}{llr}
\toprule
Checkpoint & Short name & Pretraining tokens \\
\midrule
\texttt{step330000-tokens1383B} & 1383B & 1.383T \\
\texttt{step447000-tokens1874B} & 1874B & 1.874T \\
\texttt{step512000-tokens2146B} & 2146B & 2.146T \\
\texttt{step561250-tokens2353B} & 2353B & 2.353T \\
\texttt{step600000-tokens2515B} & 2515B & 2.515T \\
\texttt{step650000-tokens2725B} & 2725B & 2.725T \\
\texttt{step738000-tokens3094B} & 3094B & 3.094T \\
\bottomrule
\end{tabular}
\end{table}

\paragraph{Instruction-tuning data.}
We use two target distributions. For Anthropic-HH, we use the
\texttt{Anthropic/hh-rlhf} training split
\citep{baiTrainingHelpfulHarmless2022} and keep the chosen transcript from each
preference pair. Each transcript is parsed into alternating user and assistant
turns, and each turn is serialized as
\begin{center}
\texttt{role: content}
\end{center}
with blank lines between turns.
For Tulu-v1, we use a deterministic 200k-example subset of the
\texttt{allenai/tulu-v1-sft-mixture} training split
\citep{wangHowFarCan2023}, selected by shuffling with seed 0 and taking the
first 200k examples. We serialize each message as \texttt{role: content},
separated by blank lines. The resulting tokenized caches contain 160,800
Anthropic-HH examples and 200,000 Tulu-v1 examples.
Training sequences are packed to length 1024. We do not use a validation split
or early stopping during fine-tuning.

\paragraph{Loss schedules and baselines.}
Let $\mathcal{L}_{\mathrm{pre}}$ denote the next-token loss on replayed
pretraining data and $\mathcal{L}_{\mathrm{inst}}$ denote the next-token loss on
the instruction-tuning target. Loss smoothing uses
\[
\mathcal{L}_t(\theta)
= (1-\alpha_t)\mathcal{L}_{\mathrm{pre}}(\theta)
+ \alpha_t \mathcal{L}_{\mathrm{inst}}(\theta),
\qquad
\alpha_t = \min(t/\tau, 1).
\]
We compare three method families. Hard-Switch sets $\alpha_t=1$ for every
fine-tuning step. Loss Smoothing linearly ramps $\alpha_t$ from 0 to 1 over
$\tau \in \{16,64,128\}$ update steps. Constant Replay is a diagnostic baseline
with fixed weights
$0.5\mathcal{L}_{\mathrm{pre}} + 0.5\mathcal{L}_{\mathrm{inst}}$ throughout
fine-tuning. The replay data is a 500M-token cap of the top OLMo pretraining
components: Common Crawl, Falcon, StarCoder, C4, Reddit, and peS2o, mixed in
proportion to their OLMo pretraining weights. Replay is included because it
also softens the pretraining-to-instruction distribution shift, but it is not
the proposed method.

\paragraph{Optimization.}
All OLMo-1B fine-tuning runs use AdamW
\citep{loshchilovDecoupledWeightDecay2018a} with $\beta_1=0.9$,
$\beta_2=0.95$, $\epsilon=10^{-8}$, global gradient clipping at norm 1.0, and
weight decay 0.
The learning-rate schedule is cosine decay to zero with 20 warmup steps. We use
bfloat16 compute with float32 parameters, global batch size 256 packed
sequences, sequence length 1024, and per-device parallelism 8. The resulting
training lengths are 141 update steps for Anthropic-HH and 483 update steps for
Tulu-v1, corresponding to 36.96M and 126.62M target tokens respectively. The
learning-rate sweep is
\[
\eta \in \{10^{-5}, 5\cdot 10^{-5}, 10^{-4}\}.
\]
For each checkpoint, target, method configuration, and learning rate, we run
data-order seeds $\{0,1,2\}$. This gives
$7$ checkpoints $\times$ $2$ targets $\times$ $3$ learning rates $\times$
$3$ seeds $\times$ $(1\ \text{hard-switch} + 1\ \text{replay} +
3\ \text{smoothing horizons}) = 630$ OLMo-1B fine-tuning runs.

\paragraph{ID evaluation.}
ID performance is an AlpacaEval-style pairwise win rate
\citep{alpaca_eval}. We generate responses to the 805
prompts from the \texttt{tatsu-lab/alpaca\_eval} eval split with the prompt
template
\begin{center}
\texttt{user: <instruction>\textbackslash n\textbackslash nassistant: }.
\end{center}
Generation uses vLLM \citep{kwonEfficientMemoryManagement2023}, bfloat16
weights, temperature 0, top-$p=1$, maximum 1024 new tokens, minimum 1 new token,
and seed 0. We judge each candidate against a target-matched hard-switch
reference trained from the 1874B checkpoint with learning rate $10^{-5}$ and
seed 0; the same reference responses are reused for all pretraining budgets
within a target family. Judging uses \texttt{Meta-Llama-3-70B-Instruct}
\citep{grattafioriLlama3Herd2024} in vLLM with tensor parallel size 4,
temperature 0, maximum 8 judge tokens, and a response character cap of 12,000.
Candidate/reference order is randomized deterministically per example. The
reported ID metric is
\[
\frac{\#\text{candidate wins} + 0.5\,\#\text{ties}}{\#\text{judged examples}},
\]
with unparseable judge outputs excluded from the denominator.

\paragraph{OOD evaluation.}
OOD performance is measured with Levanter's lm-evaluation-harness integration
on five 5-shot tasks \citep{eval-harness}: ARC-Challenge, ARC-Easy, HellaSwag,
PIQA, and WinoGrande. We evaluate the final HuggingFace export of each
fine-tuned run with maximum context length 2048 and no chat template. The OOD
summary is the mean score across these five tasks. We also retain the per-task
scores for all runs.

\section{Online Reinforcement Learning}

\subsection{Additional Results}
\begin{figure}
    \centering
    \hfil
    \begin{subfigure}{.45\textwidth}
        \includegraphics[width=\textwidth]{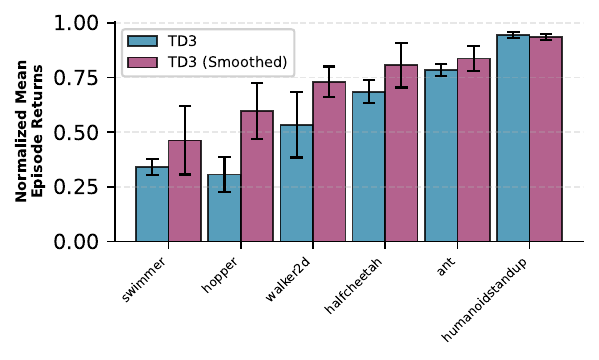}
        \caption{TD3 episode returns on Brax environments, normalized by the maximum return achieved across
        any seed of any agent.}
        \label{fig:td3_values}
    \end{subfigure}
    \hfil
    \begin{subfigure}{.5\textwidth}
        \includegraphics[width=\textwidth]{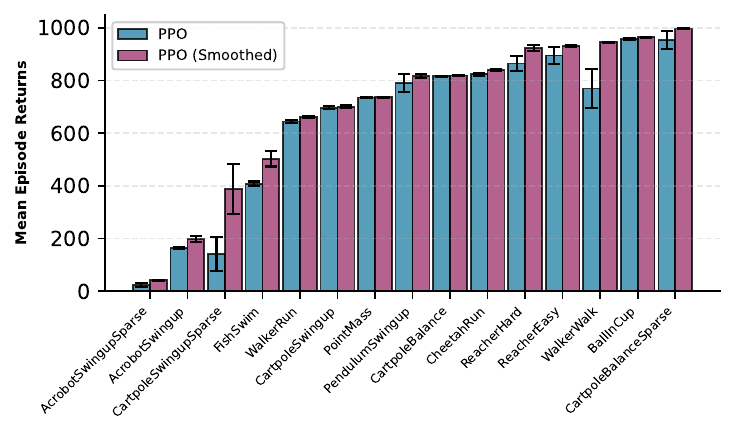}
        \caption{PPO episode returns on DeepMind Control Suite environments from Mujoco
        Playground. Maximum return for any environment is 1000.
        }
        \label{fig:ppo_values}
    \end{subfigure}
    \hfil
    \caption{TD3 and PPO final performance on each task.}
\end{figure}

We show the raw returns for the TD3 and PPO experiments in \Cref{fig:td3_values} and
\Cref{fig:ppo_values}, respectively. 

\subsection{Experimental Details}
\begin{figure}
    \centering
    \hfill
    \begin{subfigure}{.4\textwidth}
        \includegraphics[width=\textwidth]{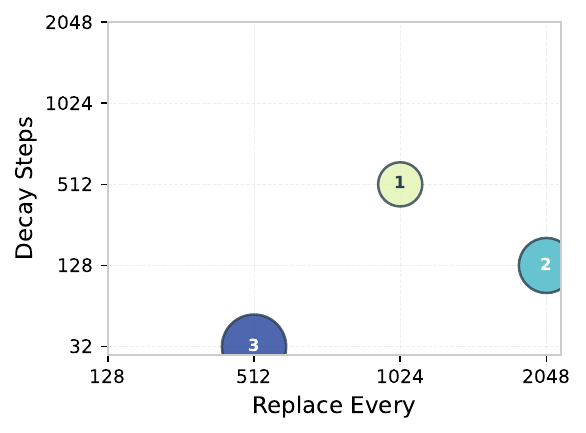}
        \caption{TD3 Smoothed Hyperparameters Selected}
        \label{fig:td3_hp}
    \end{subfigure}
    \hfill
    \begin{subfigure}{.4\textwidth}
        \includegraphics[width=\textwidth]{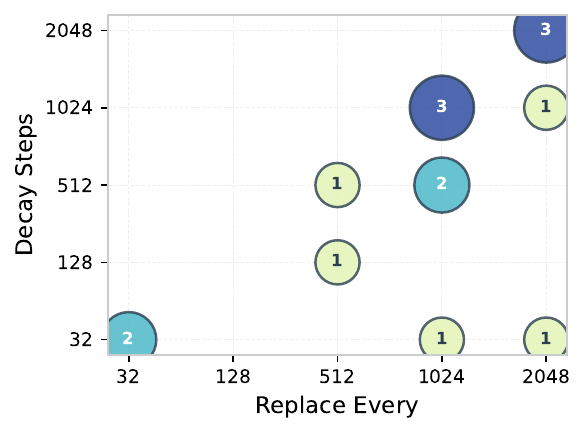}
        \caption{PPO Smoothed Hyperparameters Selected}
        \label{fig:ppo_hp}
    \end{subfigure}
    \hfill
    \caption{TD3 and PPO smoothed hyperparameters. Decay steps refers to the number of
    steps $\tau$ over which the loss is smoothed, while Replace Every refers to the
    number of steps $n$ after which the replay is updated with a new batch.
    }
\end{figure}

Our code is based on the rejax codebase \citep{rejax}.
\paragraph{TD3}
For the base algorithm, we take the hyperparameters present in the rejax codebase,
as they have been tuned for good performance on these environments. We then apply
loss smoothing to the critic loss, tuning the $\tau$ and task length hyperparameters
via grid search. Our policy and critic networks are 2 layer MLPs with 256 hidden units
and tanh activations. We train for 5 million environment steps on each task.
In \Cref{fig:td3_hp}, we present the frequency of each loss smoothing hyperparameter
configuration that we select across environments. 

\paragraph{PPO}
We train for 50 million environment steps, and use the SimbaV2 architecture
\citep{leeHypersphericalNormalizationScalable2025} as our base policy and value function
network. We tune the agent hyperparameters in stages, first tuning the base PPO
hyperparameters such as learning rate and batch size, and then tuning the loss smoothing
hyperparameters $\tau$ and $n$. In \Cref{tab:hyperparameters}, we present the
base PPO hyperparameters that we use for all experiments, and in
\Cref{fig:ppo_hp}, we present the frequency of each loss smoothing hyperparameter
configuration that we select across environments. We see that there are several
configurations that are selected quite frequently.

\begin{table}[h]
    \centering
    \begin{tabular}{lr}
        \toprule
        Hyperparameter & Value \\
        \midrule
        Max Gradient Norm & 1.0 \\
        Total Timesteps & 50,000,000 \\
        Evaluation Frequency & 512,000 \\
        Discount Factor ($\gamma$) & 0.99 \\
        GAE Lambda & 0.95 \\
        Clipping Epsilon & 0.2 \\
        Entropy Coefficient & 0.01 \\
        Value Function Coefficient & 0.5 \\
        Normalize Observations & true \\
        Number of Environments & 2048 \\
        Number of Steps & 10 \\
        Learning Rate & 0.0001 \\
        Number of Epochs & 1 \\
        Number of Minibatches & 32 \\
        Entropy Schedule & linear \\
        Starting Entropy Coefficient & 1.0 \\
        \bottomrule
    \end{tabular}
    \caption{PPO base hyperparameters}
    \label{tab:hyperparameters}
\end{table}

\section{Compute Resources Used for Main Experiments}
\label{sec:compute}
\paragraph{Continual Supervised Learning Experiments}
Each baseline for each of the two task shifts (permuted input and shuffled labels) took
about 2 hours to run on a L40S GPU (all 10 seeds). Total compute including
hyperparameter tuning took approximately 400 GPU hours.

\paragraph{Offline-to-Online RL Experiments}
AWAC and Cal-QL experiments were run on a L40S GPUs (5 seeds at once). Across all
experiments, this resulted in approximately 200 GPU hours of compute.
For the other experiments, pretraining was done on a L40S GPU (fairly quick, around 10
hours across all experiments), while online fine-tuning was done on CPU nodes with 192
CPU cores (24 seeds at once). Across all experiments and failed experiments not in the
paper, this resulted in approximately 350 CPU node hours of compute.
\paragraph{LLM Experiments}
All LLM experiments were run on L40S GPU nodes or H100 GPU nodes. We estimate that the
total compute across all experiments and evaluations was approximately 4000 GPU node hours.

\paragraph{Online RL Experiments}
Across all environments, this study took approximately 8000 GPU hours of compute on L40S GPUs.

\newpage

\end{document}